\documentclass[letterpaper]{article} 
\usepackage{aaai25}  
\usepackage{times}  
\usepackage{helvet}  
\usepackage{courier}  
\usepackage[hyphens]{url}  
\usepackage{graphicx} 
\usepackage{natbib}  
\usepackage{caption} 
\usepackage[utf8]{inputenc} 
\usepackage[T1]{fontenc}    
\usepackage{url}            
\usepackage{booktabs}       
\usepackage{amsfonts}       
\usepackage{nicefrac}       
\usepackage{microtype}      
\usepackage{xcolor}         
\usepackage{drago}
\usepackage[on]{editing}

\newcommand{\posmid}{0.5}
\newcommand{\pospart}{0.225}
\newcommand{\ttl}{Fairness-Accuracy Trade-Offs: A Causal Perspective}
\title{\ttl}

\author{
    Drago Ple\v cko,  Elias Bareinboim \\
}
\affiliations{
    Department of Computer Science\\ Columbia University \\ New York, NY 10027, United States \\
    dp3144@columbia.edu, eb@cs.columbia.edu \\
}

\usepackage{selectp}
\begin{document}

\maketitle

\begin{abstract}
  With the widespread adoption of AI systems, many of the decisions once made by humans are now delegated to automated systems. Recent works in the literature demonstrate that these automated systems, when used in socially sensitive domains, may exhibit discriminatory behavior based on sensitive characteristics such as gender, sex, religion, or race. In light of this, various notions of fairness and methods to quantify discrimination have been proposed, also leading to the development of numerous approaches for constructing fair predictors. At the same time, imposing fairness constraints may decrease the utility of the decision-maker, highlighting a tension between fairness and utility. This tension is also recognized in legal frameworks, for instance in the disparate impact doctrine of Title VII of the Civil Rights Act of 1964 -- in which specific attention is given to considerations of \textit{business necessity} -- possibly allowing the usage of proxy variables associated with the sensitive attribute in case a high-enough utility cannot be achieved without them. In this work, we analyze the tension between fairness and accuracy from a causal lens for the first time. We introduce the notion of a path-specific excess loss (PSEL) that captures how much the predictor's loss increases when a causal fairness constraint is enforced. We then show that the total excess loss (TEL), defined as the difference between the loss of predictor fair along all causal pathways vs. an unconstrained predictor, can be decomposed into a sum of more local PSELs. At the same time, enforcing a causal constraint often reduces the disparity between demographic groups. Thus, we introduce a quantity that summarizes the fairness-utility trade-off, called the causal fairness/utility ratio, defined as the ratio of the reduction in discrimination vs. the excess in the loss from constraining a causal pathway. This quantity is particularly suitable for comparing the fairness-utility trade-off across different causal pathways. Finally, as our approach requires causally-constrained fair predictors, we introduce a new neural approach for causally-constrained fair learning. Our approach is evaluated across multiple real-world datasets, providing new insights into the tension between fairness and accuracy.
\end{abstract}

\section{Introduction}
Automated decision-making systems based on machine learning and artificial intelligence are now commonly implemented in various critical sectors of society such as hiring, university admissions, law enforcement, credit assessments, and health care \citep{khandani2010consumer,mahoney2007method,brennan2009evaluating}. These technologies now significantly influence the lives of individuals and are frequently used in high-stakes settings \citep{topol2019high, berk2021fairness, taddeo2018ai}. As these systems replace or augment human decision-making processes, concerns about fairness and bias based on protected attributes such as race, gender, or religion have become a prominent consideration in the ML literature. The available data used to train automated systems may contain past and present societal biases as an imprint and therefore has the potential to perpetuate or even exacerbate discrimination against protected groups. This is highlighted by reports on biases in systems for sentencing \cite{ProPublica}, facial recognition \cite{pmlr-v81-buolamwini18a}, online ads \cite{Sweeney13,Datta15}, and system authentication \cite{Sanburn15}, among many others. Despite the promise of AI to enhance human decision-making, the reality is that these technologies can also reflect or worsen societal inequalities. As alluded to before, the issue does not arise uniquely from the usage of automated systems; human-driven decision-making has long been analyzed in a similar fashion. Evidence of bias in human decision-making is abundant, including studies on the gender wage gap \citep{blau1992gender, blau2017gender} and racial disparities in legal outcomes \citep{sweeney1992influence, pager2003mark}. Therefore, without proper care about fairness and transparency of the new generation of AI systems, it is unclear what its impact will be on the historically discriminated groups.

Within the growing literature on fair machine learning, a plethora of fairness definitions have been proposed. Commonly considered statistical criteria, among others, include demographic parity (independence \citep{darlington1971another}), equalized odds (separation \citep{hardt2016equality}), and calibration (sufficiency \citep{chouldechova2017fair}). These definitions, however, have been shown as mutually incompatible \citep{barocas2016big, kleinberg2016inherent}. Despite a number of proposals, there is still a lack of consensus on what the appropriate measures of fairness are, and how statistical notions of fairness could incorporate moral values of the society at large. For this reason, a number of works explored the causal approaches to fair machine learning \citep{kusner2017counterfactual, kilbertus2017avoiding, nabi2018fair, zhang2018fairness, zhang2018equality, wu2019pc, chiappa2019path, plevcko2020fair}, and an in-depth discussion can be found in \citep{plecko2022causal}. The main motivation for doing so is that the causal approach may allow the system designers to attribute the observed disparities between demographic groups to the causal mechanisms that underlie and generate them in the first place. In this way, by isolating disparities transmitted along different causal pathways, one obtains a more fine-grained analysis, and the capability to decide which causal pathways are deemed as unfair or discriminatory. More fundamentally, such considerations also form the basis of the legal frameworks for assessing discrimination in the United States and Europe. For instance, in the context of employment law, the disparate impact doctrine within the Title VII of the Civil Rights Act of 1964 \citep{act1964civil} disallows any form of discrimination that results in a too large of a disparity between groups of interest. A core aspect of this doctrine, however, is the notion of \textit{business necessity} (BN) or job-relatedness. Considerations of business necessity may allow variables correlated with the protected attribute to act as a proxy, and the law does not necessarily prohibit their usage due to their relevance to the business itself (or more broadly the utility of the decision-maker). Often, the wording that is used is that to argue business necessity in front of a court of law, the plaintiff needs to demonstrate that ``there is no practice that is less discriminatory and achieves the same utility'' \citep{Elston1993}. This concept illustrates the tension between fairness and utility, and demonstrates that we cannot be oblivious to considerations of utility from a legal standpoint. Moreover, demonstrating that a sufficient loss in accuracy results from imposing a fairness constraint has been previously used to justify business necessity considerations in some rulings of the European Court of Justice \citep{adams2023directly, weerts2023algorithmic}, emphasizing the importance of the topic studied in this paper.

\paragraph{Related work.} 
We mention three parts of related literature.
First, we mention the literature exploring fairness-utility trade-offs, such as  \citep{corbett2017algorithmic}. The essential argument is that an unconstrained predictor always achieves a greater or equal utility than a constrained one. Many works find that introducing fairness constraints reduces utility \citep{mitchell2021algorithmic}, and propose ways of handling the fairness-utility trade-off \citep{fish2016confidence}. 

However, other works in this literature still seem to be divided on the issue of whether trade-offs exist. For instance, \citep{rodolfa2021empirical} finds that fairness and utility trade-offs are negligible in practice, while others argue that such trade-offs need not even exist \citep{maity2020there, dutta2020there}. 
Naturally, the implications on the predictor's utility will strongly depend on the exact type of the fairness constraint that is enforced, and works that do not find a trade-off often focus on equality of odds \citep{hardt2016equality} ($\widehat Y \ci X \mid Y$) or (multi)calibration \citep{chouldechova2017fair} ($Y \ci X \mid \widehat Y$). Notably, the former metric always allows for the perfect predictor $\widehat Y = Y$, and thus in settings with good predictive power, the cost of enforcing this constraint may indeed be negligible. The latter metric allows for the $L_2$-optimal prediction score, and improving miscalibration may sometimes also yield improvements in utility by serving as a type of regularization.

Finally, in the causal fairness literature, the tension between fairness and utility has been largely unexplored. Some exceptions include works such as \citep{nilforoshan2022causal} which shows that for a decision policy that satisfies a causal fairness constraints, it is almost always possible to find another decision that has a higher utility and the same total variation (TV) measure. \cite{plecko2024causal} performs a causal explanation on a decision score used for constructing a policy, and discusses how disparities in the decision score may influence utility. Our main aim of this paper is to fill in this gap, and provide a systematic way of analyzing the fairness-accuracy trade-off from a causal lens, and show that fairness and utility are almost always in a trade-off.

\subsection{Motivating Example}
We illustrate our approach in a simple linear setting:
\begin{example}[Linear Fairness-Accuracy Causal Trade-Offs] \label{ex:lfacto}
    Consider variables $X, W, Y$ behaving according to the following linear system of equations:
    \begin{align}
        X &\gets \text{Bernoulli}(0.5) \label{eq:lfacto-scm-1}\\
        W &\gets \beta X + \epsilon_w \\
        Y &\gets \alpha X + \gamma W + \epsilon_y, \label{eq:lfacto-scm-3}
    \end{align}
    where $\epsilon_w \sim N(0, \sigma_w^2), \epsilon_y \sim N(0, \sigma_y^2)$. Variable $X$ is the protected attribute, and $Y$ is the outcome of interest. The causal diagram of Eqs.~\ref{eq:lfacto-scm-1}-\ref{eq:lfacto-scm-3} is shown in Fig.~\ref{fig:sfm} (with the $Z$ set empty). 
    Attribute $X$ can influence $Y$ along two different pathways: the direct path $X \to Y$, and the indirect path $X \to W \to Y$. Therefore, for considering fairness, we focus on fair linear predictors $\widehat Y^S$ of the form
    \begin{align}
        \widehat Y ^ S = \hat \alpha_S X + \hat \gamma_S W,
    \end{align}
    where the predictor $\widehat Y^S$ removes effects in the set $S$, with $S$ ranging in $\{ \emptyset, \text{DE}, \text{IE}, \{ \text{DE, IE} \}\}$ (DE, IE stand for direct and indirect effects, and any subset of these could be removed). For instance, the optimal predictor $\widehat Y ^  \emptyset$, which is not subject to any fairness constraints, has the coefficients 
    \begin{align}
        \hat \alpha_{\emptyset} = \alpha, \hat \gamma_{\emptyset} = \gamma,
    \end{align}
    which are the ordinary least squares (OLS) coefficients. Therefore, its mean-squared error (MSE) can be computed as $\ex [Y - \widehat Y ^  \emptyset]^2 = \sigma_y^2$. 
    The DE-fair predictor $\widehat Y^{\text{DE}}$, which has the direct effect constrained to zero, has coefficients
    \begin{align}
        \hat\alpha_{\text{DE}} = 0, \hat\gamma_{\text{DE}} = \gamma.
    \end{align}
    The fully-fair predictor, labeled $\widehat Y^{\{ \text{DE, IE} \}}$, has both direct and indirect effects constrained to zero, and thus has coefficients
    \begin{align}
        \hat\alpha_{\{ \text{DE, IE} \}} = 0, \hat\gamma_{\{ \text{DE, IE} \}} = 0.
    \end{align}
    Thus, the corresponding MSE values for $\widehat Y^{\text{DE}}$ and $\widehat Y^{\{ \text{DE, IE} \}}$ can be computed as (see Appendix~\ref{appendix:linear-psel} for computation details):
    \begin{align}
        \ex [Y - \widehat Y^\text{DE}]^2 &=  \sigma_y^2 + \frac{\alpha^2}{2}, \\
        \hspace{-6pt}\ex [Y - \widehat Y^{\{ \text{DE, IE} \}}]^2 &=  \sigma_y^2 + \frac{\alpha^2 + \gamma^2 \beta^2 + \alpha\gamma\beta}{2} + \gamma^2 \sigma_w^2.
    \end{align}
    Our goal is to decompose the total excess loss (TEL) originating from imposing the fairness constraints, defined as:
    \begin{align}
        \text{TEL} := \underbrace{\ex [Y - \widehat Y^{\{ \text{DE, IE} \}}]^2}_{\text{fully-fair predictor's loss}} - \underbrace{\ex [Y - \widehat Y ^  \emptyset]^2}_{\text{unconstrained loss}}.
    \end{align}
    TEL measures the excess loss (in terms of the increase in the MSE, compared to the unconstrained predictor) originating from the removal of the direct and indirect effects. This quantifies the excess loss traded off for enforcing fairness constraints. Our goal is to decompose the TEL to obtain path-specific contributions originating from the removal of direct and indirect effects as follows:
    \begin{align}
         \text{TEL}
         &=\underbrace{\ex [Y - \widehat Y^{\text{DE}}]^2 - \ex [Y - \widehat Y ^  \emptyset]^2}_{\text{Term I = excess DE loss}} \label{eq:de-psel} \\
         &+ \underbrace{\ex [Y - \widehat Y^{\{ \text{DE, IE} \}}]^2 - \ex [Y - \widehat Y ^ {\text{DE}}]^2}_{\text{Term II = excess IE loss}} \label{eq:ie-psel} \\
         &= \underbrace{\frac{\alpha^2}{2}}_{\text{Term I}} + \underbrace{\frac{\gamma^2 \beta^2 + \alpha\gamma\beta}{2} + \gamma^2 \sigma_w^2}_{\text{Term II}}.
    \end{align}
    Term I is the direct effect excess loss, incurred by constraining the direct effect to 0. Term II is the indirect effect excess loss, incurred by constraining the indirect effect to 0.

\begin{figure}
        \centering
        \includegraphics[width=\linewidth]{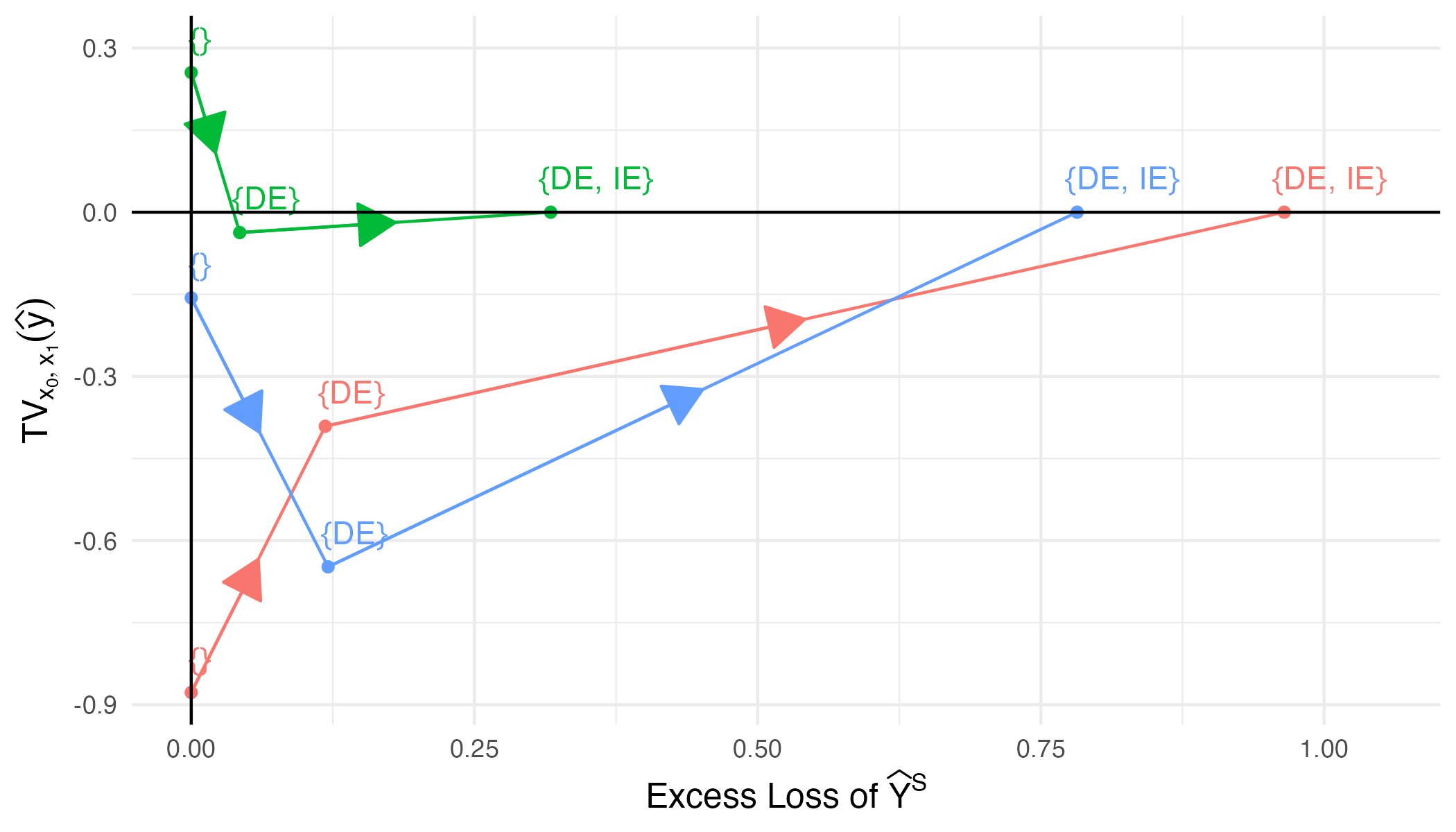}
        \caption{Total Variation (TV) measures vs. Excess Loss. Different colors represent trajectories obtained for different randomly sampled linear SCMs.}
        \label{fig:intro-plot}
\end{figure}

    At the same time, enforcing fairness constraints may reduce the disparity between groups. For any predictor $\widehat Y$, we can measure the disparity using the difference in conditional expectations, called the total variation measure, defined as $\text{TV}_{x_0, x_1}(\widehat y) = \ex[\widehat Y \mid x_1] - \ex[\widehat Y \mid x_0]$. This measure is sometimes also referred to as the parity gap. 
    Similarly as for the TEL, we compute the decrease in group disparity associated with removing DE and IE, by comparing the TV measures of the fully-fair $\widehat Y ^{\{\text{DE, IE}\}}$ and unconstrained $\widehat Y ^ \emptyset$, and computing the TV difference (TVD, for short), defined as:
    \begin{align}
        \text{TVD} &=\text{TV}(\widehat Y ^{\{\text{DE, IE}\}}) - \text{TV} (\widehat Y ^ \emptyset) \\
        &= \underbrace{(\ex[\widehat Y ^{\{\text{DE, IE}\}} \mid x_1] - \ex[\widehat Y ^{\{\text{DE, IE}\}} \mid x_0])}_{\text{disparity after removing DE, IE}} \\ & \quad - \underbrace{(\ex[\widehat Y ^ \emptyset \mid x_1] - \ex[\widehat Y ^ \emptyset \mid x_0])}_{\text{disparity before removing DE, IE}} = -\alpha - \beta\gamma.
    \end{align}
    The TVD metric again decomposes into the contributions along direct and indirect effects:
    \begin{align}
        \text{TVD} &= \underbrace{\text{TV}(\widehat Y ^{\text{DE}}) - \text{TV} (\widehat Y ^ \emptyset)}_{\text{Term A = disparity reduction of DE}} \\
        &\quad +
        \underbrace{\text{TV}(\widehat Y ^{\{\text{DE, IE}\}}) - \text{TV} (\widehat Y ^ {\text{DE}})}_{\text{Term B = disparity reduction of IE}}.
    \end{align}
    Terms A and B can be computed as:
    \begin{align} \label{eq:de-tvd}
        \text{Term A} = &(\ex[\widehat Y ^ \text{DE} \mid x_0] - \ex[\widehat Y ^ \text{DE} \mid x_0]) \\ &- (\ex[\widehat Y ^ \emptyset \mid x_0] - \ex[\widehat Y ^ \emptyset \mid x_0]) = -\alpha \\
        \text{Term B} = &(\ex[\widehat Y ^{\{\text{DE, IE}\}} \mid x_1] - \ex[\widehat Y ^{\{\text{DE, IE}\}} \mid x_0]) \\ &- (\ex[\widehat Y ^ \text{DE} \mid x_0] - \ex[\widehat Y ^ \text{DE} \mid x_0]) = -\beta\gamma. \label{eq:ie-tvd}
    \end{align}
    Based on the excess loss and the reduction in disparity resulting from constraining a causal path to zero, we can quantify the fairness/utility trade-off through a causal lens. 
    Prototypical instances for predictors $\widehat Y^S$ are visualized in Fig.~\ref{fig:intro-plot}, for three randomly drawn triples $(\alpha, \beta, \gamma)$. Predictor $\widehat Y^\emptyset$ is optimal and thus always has $0$ excess loss (and hence lies on the vertical axis). Fully-fair predictor $\widehat Y^{\{\text{DE, IE}\}}$ removes both direct and indirect effects, and thus always has the TV measure equal to $0$.
    Therefore, $\widehat Y^{\{\text{DE, IE}\}}$ always lies on the horizontal axis (corresponding to TV = 0). The slopes in the plot between $\widehat Y^\emptyset$, $\widehat Y ^ \text{DE}$, and $\widehat Y ^ \text{DE, IE}$ (indicated by arrows), geometrically capture the tension between excess loss and reducing discrimination upon imposing a constraint. These slopes are computed as the ratio
    \begin{align}
        \frac{\text{TV difference}}{\text{Excess Loss}},
    \end{align}
    a quantity that we call the causal fairness-utility ratio (CFUR). Based on Eqs.~\ref{eq:de-psel}-\ref{eq:ie-psel} and \ref{eq:de-tvd}-\ref{eq:ie-tvd} we can compute
    \begin{align}
        \text{CFUR}&(\text{DE}) = -\frac{1}{\alpha}, \\
        \text{CFUR}(\text{IE}) =& -\frac{2\beta\gamma}{\gamma^2 \beta^2 + \alpha\gamma\beta + 2\gamma^2\sigma_w^2},
    \end{align}
    summarizing the fairness-utility trade-off for each path.
\end{example}

\noindent The above example illustrates how in the simple linear case we can attribute the increased loss from imposing fairness constraints to the specific causal pathway in question. It also shows we can compute the associated change in the disparity between groups, quantified by the TV measure $\ex[\widehat Y \mid x_1]-\ex[\widehat Y \mid x_0]$. In this paper, we generalize the approach from the above example to a non-parametric setting, with the following key contributions:
\begin{enumerate}[label=(\roman*)]
    \item We introduce the notion of a \textit{path-specific excess loss} (PSEL) associated with imposing a fairness constraint along a causal path (Def.~\ref{def:psel}), and we prove how the total excess loss (TEL) can be decomposed into a sum of path-specific excess losses (Thm.~\ref{thm:tpl-decomposition}),
    \item We develop an algorithm for attributing path-specific excess losses to different causal paths (Alg.~\ref{algo:pspl-attribution}), allowing the system designer to explain how the total excess loss is affected by different fairness constraints. In this context, we show the equivalence of Alg.~\ref{algo:pspl-attribution} with a Shapley value \citep{shapley1953value} approach (Prop~\ref{prop:pspl-shapley}),
    \item For purposes of applying Alg.~\ref{algo:pspl-attribution}, a key requirement is the construction of causally-fair predictors $\widehat Y^S$ that remove effects along pathways in $S$. We introduce a novel Lagrangian formulation of the optimization problem for such $\widehat Y^S$ (Def.~\ref{def:lagrange-form}) accompanied with a training procedure for learning the predictor (Alg.~\ref{algo:lagrange-training}),
    \item We introduce the causal fairness/utility ratio (CFUR, Def.~\ref{def:cfur}) that summarizes how much the group disparity can be reduced per fixed cost in terms of excess loss. We compute CFURs on a range of real-world datasets, and demonstrate that from a causal viewpoint fairness and utility are almost always in tension.
\end{enumerate}

\subsection{Preliminaries} \label{sec:prelims}
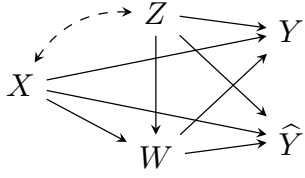
\begin{figure}
        \centering
	\scalebox{1.2}{
        \begin{tikzpicture}
	 [>=stealth, rv/.style={thick}, rvc/.style={triangle, draw, thick, minimum size=7mm}, node distance=18mm]
	 \pgfsetarrows{latex-latex};
	 \begin{scope}
		\node[rv] (0) at (0,0.8) {$Z$};
	 	\node[rv] (1) at (-1.5,0) {$X$};
	 	\node[rv] (2) at (0,-0.8) {$W$};
	 	\node[rv] (3) at (1.5,0.6) {$Y$};
            \node[rv] (4) at (1.5,-0.6) {$\widehat{Y}$};
	 	\draw[->] (1) -- (2);
		\draw[->] (0) -- (3);
            \draw[->] (0) -- (4);
	 	\path[->] (1) edge[bend left = 0] (3);
            \path[->] (1) edge[bend left = 0] (4);
		\path[<->] (1) edge[bend left = 30, dashed] (0);
	 	\draw[->] (2) -- (3);
            \draw[->] (2) -- (4);
		\draw[->] (0) -- (2);
	 \end{scope}
	 \end{tikzpicture}
        }
    \caption{Standard Fairness Model, with the protected attribute $X$, set of confounders $Z$, set of mediators $W$, outcome $Y$, and a predictor $\widehat Y$.}
    \label{fig:sfm}
\end{figure}
We use the language of structural causal models (SCMs) \citep{pearl:2k}. An SCM is
a tuple $\mathcal{M} := \langle V, U, \mathcal{F}, P(u)\rangle$ , where $V$, $U$ are sets of
endogenous (observable) and exogenous (latent) variables, 
respectively, $\mathcal{F}$ is a set of functions $f_{V_i}$,
one for each $V_i \in V$, where $V_i \gets f_{V_i}(\pa(V_i), U_{V_i})$ for some $\pa(V_i)\subseteq V$ and
$U_{V_i} \subseteq U$. $P(u)$ is a strictly positive probability measure over $U$. Each SCM $\mathcal{M}$ is associated to a causal diagram $\mathcal{G}$ \citep{bareinboim2020on} over the node set $V$ where $V_i \rightarrow V_j$ if $V_i$ is an argument of $f_{V_j}$, and $V_i \bidir V_j$ if the corresponding $U_{V_i}, U_{V_j}$ are not independent. An instantiation of the exogenous variables $U = u$ is called a \textit{unit}. By $Y_{x}(u)$ we denote the potential response of $Y$ when setting $X=x$ for the unit $u$, which is the solution for $Y(u)$ to the set of equations obtained by evaluating the unit $u$ in the submodel $\mathcal{M}_x$, in which all equations in $\mathcal{F}$ associated with $X$ are replaced by $X = x$. 
For more details about the causal inference background, we refer the reader to \citep{pearl:2k, bareinboim2020on, plecko2022causal}.
Throughout the paper, we assume a specific cluster causal diagram $\mathcal{G}_{\text{SFM}}$ known as the standard fairness model (SFM) \citep{plecko2022causal} over endogenous variables $\{X, Z, W, Y, \widehat{Y}\}$ shown in Fig.~\ref{fig:sfm}. The SFM consists of the following: \textit{protected attribute}, labeled $X$ (e.g., gender, race, religion), assumed to be binary; the set of \textit{confounding} variables $Z$, which are not causally influenced by the attribute $X$ (e.g., demographic information, zip code); the set of \textit{mediator} variables $W$ that are possibly causally influenced by the attribute (e.g., educational level or other job-related information); the \textit{outcome} variable $Y$ (e.g., GPA, salary); the \textit{predictor} of the outcome $\widehat{Y}$ (e.g., predicted GPA, predicted salary). The SFM encodes the assumptions typically used in the causal inference literature about the lack of hidden confounding. The availability of the SFM and the implied assumptions are a possible limitation of the paper, while we note that partial identification techniques for bounding effects can be used for relaxing them \citep{zhang2022partialctf}.
Based on the SFM, we will use the following causal fairness measures:
\begin{definition}[Population-level Causal Fairness Measures \citep{pearl:01, plecko2022causal}]
    The natural direct, indirect, and spurious effects are defined as
    \begin{align}
        \text{NDE}_{x_0, x_1}(y) &= P(y_{x_1, W_{x_0}}) - P(y_{x_0}) \label{eq:nde}  
        \\
        \text{NIE}_{x_1, x_0}(y) &= P(y_{x_1, W_{x_0}}) - P(y_{x_1}) \label{eq:nie}  
        \\
        \text{NSE}_{x}(y) &= P(y \mid x) - P(y_x). \label{eq:nse}
    \end{align}
\end{definition}
The NDE in Eq.~\ref{eq:nde} compares the potential outcome $Y_{x_1, W_{x_0}}$, in which $Y$ responds to $X = x_1$ along the direct path, while $W$ is set at the value it would attain naturally when responding to $X = x_0$, against the potential outcome $Y_{x_0}$ where $X = x_0$ along both direct and indirect paths. In this way, the NDE measures the variations induced by changing $x_0 \to x_1$ along the direct causal path, quantifying direct discrimination. Similarly, the NIE in Eq.~\ref{eq:nie} compares $Y_{x_1, W_{x_0}}$ vs. $Y_{x_1}$, and thus captures variations induced by considering a change $x_1 \to x_0$ along the indirect causal path (note that both $Y_{x_1, W_{x_0}}$ and $Y_{x_1}$ respond to $X = x_1$ along the direct path, so only indirect variations are induced when taking the difference). Finally, the NSE in Eq.~\ref{eq:nse} compares $Y \mid X = x$ vs. $Y_{x}$. In the former, due to conditioning on $X = x$, the distribution over the set of confounders $Z$ (in Fig.~\ref{fig:sfm}) changes according to this conditioning, while in the potential outcome $Y_{x}$, the distribution of $Z$ does not change, since $X = x$ is set by intervention. Therefore, taking the difference captures the spurious effect of $X$ on $Y$, along the backdoor path $X \bidir Z \to Y$. 
A causally-fair predictor for a subset $S$ of the above measures is defined as:
\begin{definition}[Causally Fair Predictor \citep{plecko2024reconciling}] \label{def:cfair-predictor}
    The optimal causally $S$-fair predictor $\widehat Y ^ S$ with respect to a loss function $L$ and pathways in $S$ is the solution to the following optimization problem:
    \begin{alignat}{2}
    \label{eq:inproc-causal-objective}
    &\argmin_{f}        &\quad \ex \; &L(Y, f(X, Z, W)) \\
    &\quad\text{s.t.} & \;     \text{NDE}_{x_0, x_1}(f) &= \text{NDE}_{x_0, x_1}(y) \cdot \mathbb{1}(\text{DE} \notin S)\\
    &&       \text{NIE}_{x_1, x_0}(f) &= \text{NIE}_{x_1, x_0}(y) \cdot \mathbb{1}(\text{IE} \notin S)\\
    &&       \text{NSE}_{x_0}(f) &= \text{NSE}_{x_0}(y) \cdot \mathbb{1}(\text{SE} \notin S)\\
    &&       \text{NSE}_{x_1}(f) &= \text{NSE}_{x_1}(y) \cdot \mathbb{1}(\text{SE} \notin S).
\end{alignat}
\end{definition}
The definition of $\widehat{Y}^S$ has a straightforward interpretation. For any pathway in the set $S$, the corresponding causal effect should be $0$, as proposed in the path-specific causal fairness literature \citep{nabi2018fair, chiappa2019path}. However, importantly, pathways that are not in $S$ also need to be constrained -- the effect of $X$ on $\widehat Y$ along these paths should not change compared to the true outcome $Y$ \citep{plecko2024reconciling}. 
For instance, if the direct path is not in $S$ (meaning it is considered to be non-discriminatory), then we expect to have $\text{NDE}_{x_0, x_1}(\widehat{y}) = \text{NDE}_{x_0, x_1}(y)$ (and similarly for IE, SE). This form of constraint on discriminatory pathways ensures that no undesirable bias amplification occurs along non-discriminatory pathways.
\section{Path-Specific Excess Loss} \label{sec:psel}
In this section, we introduce the concept of a path-specific excess loss, and then demonstrate how the total excess loss can be decomposed into path-specific excess losses.
\begin{definition}[Path-Specific Excess Loss] \label{def:psel}
    Let $L(\widehat Y, Y)$ be a loss function and $\widehat Y ^S$ the optimal causally $S$-fair predictor with respect to $L$.  Define the path-specific excess loss (PSEL) of a pair $S, S'$ as:
    \begin{align}
        \text{PSEL} (S \to S') = \ex [ L(\widehat Y ^ {S'}, Y) ] - \ex [ L(\widehat Y ^ {S}, Y) ].
    \end{align}
    The quantity $\text{PSEL} (\emptyset \to \{D, I, S\})$ is called the total excess loss (TEL).
\end{definition}
The total excess loss computes the increase in the loss for the totally constrained predictor $\widehat Y^{\{D, I, S\}}$ with direct, indirect, and spurious effects removed compared to the unconstrained predictor $\widehat Y^ \emptyset$. \footnote{Other classifiers that remove only subsets of the causal paths between $X$ and $Y$ may be considered fair, depending on considerations of business necessity \citep{plecko2022causal}. The rationale developed in this paper can be adapted to such settings.} 
In the following theorem, we show that the total excess loss can be decomposed as a sum of path-specific excess losse. All proofs are provided in Appendix~\ref{appendix:theorem-proofs} (for supplements, see full paper version at \url{https://arxiv.org/abs/2405.15443}):  
\begin{theorem}[Total Excess Loss Decomposition] \label{thm:tpl-decomposition}
    The total excess loss $\text{PSEL} (\emptyset \to \{D, I, S\})$ can be decomposed into a sum of path-specific excess losses as follows:
    \begin{align}
        \hspace{-8pt}\text{PSEL} (\emptyset \to \{D, I, S\}) &= \text{PSEL} (\emptyset \to \{ D \}) \\ & + \text{PSEL} (\{D\} \to \{D, I\}) \\ &+ \text{PSEL} (\{D, I \} \to \{D, I, S\}).
    \end{align}
\end{theorem}
\begin{remark}[Non-Uniqueness of Decomposition] \label{remark:non-unique-decomp}
    The decomposition in Thm.~\ref{thm:tpl-decomposition} is not unique. In particular, the $\text{PSEL} (\emptyset \to \{D, I, S\})$ can be decomposed as
    \begin{align} \label{eq:tel-decompos}
        & \text{PSEL} (\emptyset \to \{ S_1 \}) + \text{PSEL} (\{S_1\} \to \{S_1, S_2\}) \\& \quad + \text{PSEL} (\{S_1, S_2 \} \to \{D, I, S\})
    \end{align}
    for any choice of $S_1, S_2 \in \{D, I, S\}$ with $S_1 \neq S_2$. Therefore, six different decompositions exist (three choices for $S_1$, two for $S_2$). 
\end{remark}
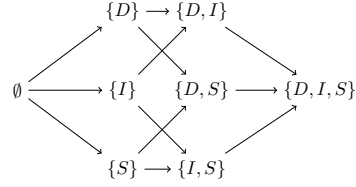
\begin{figure}
    \centering
    \scalebox{0.85}{
    \begin{tikzpicture}[node distance=1.5cm]
\node (empty) {$\emptyset$};
\node[right of=empty, xshift=0.5cm] (I) {$\{I\}$};
\node[above of=I] (D) {$\{D\}$};
\node[below of=I] (S) {$\{S\}$};
\node[right of=D] (DI) {$\{D, I\}$};
\node[right of=S] (IS) {$\{I, S\}$};
\node[right of=I] (DS) {$\{D, S\}$};
\node[right of=DS, xshift=0.75cm] (DIS) {$\{D, I, S\}$};

\draw[->] (empty) -- (D) node[sloped, above, pos=\posmid] {};
\draw[->] (empty) -- (I) node[sloped, above, pos=\posmid] {};
\draw[->] (empty) -- (S) node[sloped, above, pos=\posmid] {};

\draw[->] (S) -- (IS) node[sloped, above, pos=\posmid] {};
\draw[->] (D) -- (DI) node[sloped, above, pos=\posmid] {};

\draw[->] (I) -- (DI) node[sloped, above, pos=\pospart] {};
\draw[->] (I) -- (IS) node[sloped, above, pos=\pospart] {};
\draw[->] (D) -- (DS) node[sloped, above, pos=\pospart] {};
\draw[->] (S) -- (DS) node[sloped, above, pos=\pospart] {};

\draw[->] (DI) -- (DIS) node[sloped, above, pos=\posmid] {};
\draw[->] (IS) -- (DIS) node[sloped, above, pos=\posmid] {};
\draw[->] (DS) -- (DIS) node[sloped, above, pos=\posmid] {};
\end{tikzpicture}
    }
\caption{Graphical representation $\gpsel$.}
    \label{fig:pspl-overview}
\end{figure}
Fig.~\ref{fig:pspl-overview} provides a graphical overview of all the possible path-specific excess loses. In the left side, we start with $S=\emptyset$ and the predictor $\widehat Y ^ \emptyset$. Then, we can add any of $\{ D, I, S \}$ to the $S$-set, to obtain the predictors $\widehat Y ^ D, \widehat Y ^ I$, or $\widehat Y ^ S$, and so on. The graph representing all the possible states $\widehat Y^S$ and transitions between pairs $(\widehat Y^ S, \widehat Y^{S'})$ shown in Fig.~\ref{fig:pspl-overview} is labeled $\gpsel$. 
There are six paths starting from $\emptyset$ and ending in $\{ D, I , S \}$. In Alg.~\ref{algo:pspl-attribution}, we introduce a procedure that sweeps over all the edges and paths in $\gpsel$ to compute path-specific excess losses, while also computing the change in the TV measure between groups in order to track the reduction in discrimination. In the main body of the paper we discuss fairness-utility trade-offs when considering direct, indirect, and spurious effects. In Appendix~\ref{appendix:path-specific} we described how to adapt all the results to a general setting considering more granular path-specific effects.
\begin{algorithm}[t]
\caption{Path-Specific Excess Loss Attributions}
\label{algo:pspl-attribution}
\DontPrintSemicolon
\KwIn{data $\mathcal{D}$, predictors $\widehat{Y} ^ S$ for $S$-sets $\subseteq \{ D, I , S\}$}
\ForEach{edge $(S, S') \in \gpsel$} {
    compute the path-specific excess loss of $S' \setminus S$, given by $\text{PSEL}(S \to S')$\;
    compute the TV measure difference of $S' \setminus S$, written $\text{TVD}(S \to S')$ given by $\text{TV}_{x_0, x_1}(\widehat Y ^{S'}) - \text{TV}_{x_0, x_1}(\widehat Y ^S)$\;
}
\ForEach{causal path $S_i$ $\in \{D, I, S \}$} {
    compute the average path-specific excess loss and TV difference across all paths $\emptyset \to \dots \to \{D, I, S\}$ in $\gpsel$ \;
    $\text{APSEL}(S_i)$ is computed as
    \begin{align}
         \frac{1}{3!} \sum_{\substack{\pi \in \gpsel: \\ \emptyset \text{ to } \{D, I, S\}}} \text{PSEL}(\pi^{<S_i}  \to \pi^{<S_i} \cup S_i) \label{eq:apsel-compute}
    \end{align} \;
    $\text{ATVD}(S_i)$ is computed as
    \begin{align}
         \frac{1}{3!} \sum_{\substack{\pi \in \gpsel: \\ \emptyset \text{ to } \{D, I, S\}}} \text{TVD}(\pi^{<S_i}  \to \pi^{<S_i} \cup S_i), \label{eq:atvd-compute}
    \end{align}
    where $\pi^{<S_i}$ denotes the set of causal paths that precede $S_i$ in $\pi$.
}
\Return{set of $\text{PSEL}(S \to S')$, $\text{TVD}(S \to S')$, attributions APSEL($S_i$), ATVD($S_i$)}
\end{algorithm}
Formally, for any edge $(S, S')$ in $\gpsel$ the value of $\text{PSEL}(S \to S')$ is computed, together with the difference in the TV measure (TVD) from the transition $S \to S'$, defined by
\begin{align}
    \text{TVD}(S \to S') =& \underbrace{\ex[Y^{S'} \mid x_1] - \ex[Y^{S'} \mid x_0]}_{\text{TV after removing } S'\setminus S} \\ & - \underbrace{\ex[Y^{S} \mid x_1] - \ex[Y^{S} \mid x_0]}_{\text{TV before removing } S'\setminus S}.
\end{align}
The quantities $\text{PSEL}(S \to S')$ and $\text{TVD}(S \to S')$ are naturally associated with the effect that was removed, i.e., $S' \setminus S$. 
In this context, we mention a connection with previous work \citep{zhang2018fairness}, which provides a way of quantifying direct, indirect, and spurious effects of the attribute $X$ on the outcome $Y$. In Appendix~\ref{appendix:causal-decomposition}, we show that our approach of quantifying the change in the TV measure through the TVD quantity in practice closely corresponds the approach proposed by \citep{zhang2018fairness}.

As there are multiple ways of reaching the set $\{ D, I , S \}$ from $\emptyset$ in $\gpsel$, each of the corresponding effects (direct, indirect, spurious) will be associated with a number of different PSELs and TVDs (generally, note that the complexity is exponential in the number of causal paths included). In Eqs.~\ref{eq:apsel-compute}-\ref{eq:atvd-compute} inside the algorithm, we compute the average PSEL and TVD across all the edges that are associated with a specific effect $S_i$. This simple intuition, corresponding to taking an average across all of the possible decompositions of the total excess loss (Eq.~\ref{eq:tel-decompos}), turns out to be equivalent to a Shapley value \citep{shapley1953value} of a suitably chosen value function:
\begin{proposition}[PSEL Attribution as Shapley Values] \label{prop:pspl-shapley}
    Let the functions $f_1(S), f_2(S)$ be defined as:
    \begin{align}
        f_1(S) &= \text{PSEL} ( \emptyset \to S) \\ 
        f_2(S) &= \text{TVD} ( \emptyset \to S).
    \end{align}
    The Shapley value $\phi^k (S_i)$ for the effect $S_i \in \{D, I, S \}$ and function $f_k$, is computed as
    \begin{align}
         \sum_{S \subseteq \{D, I, S \} \setminus \{S_i\}} \frac{1}{n\binom{n-1}{|S|}} \left(f_k(S \cup \{S_i\}) - f_k(S)\right).
    \end{align}
    where $n=3$ for the choice $\{D, I, S\}$. 
    The averaged path-specific excess loss of $S_i$ and the averaged TV difference of $S_i$ are equal to the Shapley values of $S_i$ associated with functions $f_1, f_2$, respectively,
    \begin{align}
        \phi^1(S_i) &= \text{APSEL}(S_i) \\ 
        \phi^2(S_i) &= \text{ATVD}(S_i),
    \end{align}
    with $\text{APSEL}(S_i)$ defined as
    \begin{align}
         \frac{1}{3!} \sum_{\substack{\pi \in \gpsel: \\ \emptyset \text{ to } \{D, I, S\}}} \text{PSEL}(\pi^{<S_i}  \to \pi^{<S_i} \cup S_i) \label{eq:apsel-compute-inline},
    \end{align}
    and $\text{ATVD}(S_i)$ defined as
    \begin{align}
         \frac{1}{3!} \sum_{\substack{\pi \in \gpsel: \\ \emptyset \text{ to } \{D, I, S\}}} \text{TVD}(\pi^{<S_i}  \to \pi^{<S_i} \cup S_i). \label{eq:atvd-compute-inline}
    \end{align}
\end{proposition}
The above proposition illustrates how averaging the influence of removing a causal effect over all possible ways of reaching $\{D, I, S\}$ from $\emptyset$ is equivalent to computing the Shapley values of an appropriate value function $f$. 
We remark that the attribution in Prop.~\ref{prop:pspl-shapley} is not the only approach one could take. An alternative would be to average the contributions of each pathways across \textit{all edges} of the graph $\gpsel$ (instead of focusing on all pathways from $\emptyset$ to $\{D, I, S\}$). Such an attribution would not satisfy the Shapley axioms, however. 
We next introduce the notion of a causal fairness/utility ratios:
\begin{definition}[Causal Fairness/Utility Ratio (CFUR)] \label{def:cfur}
    The causal fairness/utility ratio (CFUR) for a causal path $S_i$ is defined as
    \begin{align} \label{eq:cfur}
        \text{CFUR}(S_i) = \frac{\text{ATVD}(S_i)}{\text{APSEL}(S_i)}.
    \end{align}
    Whenever $\text{APSEL}(S_i) = 0$,  $\text{CFUR}(S_i)$ is equal to $0$.
\end{definition}
The CFUR quantity may be particularly useful for comparing different causal effects, and the connection of the CFUR with local TVD and PSEL values is described in Appendix~\ref{appendix:cfur-weighted-average}. The intuition behind the quantity is simple -- for removing a causal effect $S_i$ from our predictor $\widehat Y$, we want to compute how much of a reduction in the disparity that results in (measured in terms of the ATVD measure) \textit{per unit change in the incurred excess loss}. This quantity attempts to assign a single number to a causal path that succinctly summarizes how much fairness is gained vs. how much predictive power is lost by imposing such a causal constraint. 

\section{Causally-Fair Constrained Learning}
In the preceding section, we developed an approach for quantifying the tension between fairness and accuracy from a causal viewpoint. The results were contingent on finding the optimal causally-fair predictors $\widehat Y ^S$ following Def.~\ref{def:cfair-predictor}. However, computing the predictors $\widehat Y ^ S$ is quite challenging in practice, due to several complex causal constraints in the optimization problem. We now develop a practical approach for solving this problem, by first introducing a Lagrangian form of the optimal causally-fair predictor:
\begin{definition}[Causal Lagrange Predictor $\widehat Y ^ S$] \label{def:lagrange-form}
    The causally $S$-fair $\lambda$-optimal predictor $\widehat Y ^ S(\lambda)$ with respect to pathways in $S$ and the loss function $L$ is the solution of the following:
    {
    \small
    \begin{align} \label{eq:lagrange-obj-1}
        \argmin_{f} & \; \ex \; L(Y, f(X, Z, W)) + 
        \\& 
        \lambda \big(\text{NDE}_{x_0, x_1}(f) - \text{NDE}_{x_0, x_1}(y) \cdot \mathbb{1}(\text{DE} \notin S)\big)^2 + 
        \\[-6pt]& 
        \lambda \big(\text{NIE}_{x_1, x_0}(f) - \text{NIE}_{x_1, x_0}(y) \cdot \mathbb{1}(\text{IE} \notin S)\big)^2 + 
        \\& 
        \lambda \big(\text{NSE}_{x_0}(f) - \text{NSE}_{x_0}(y) \cdot \mathbb{1}(\text{SE} \notin S)\big)^2 + 
        \\& 
        \lambda \big(\text{NSE}_{x_1}(f) - \text{NSE}_{x_1}(y) \cdot \mathbb{1}(\text{SE} \notin S)\big)^2 \label{eq:lagrange-obj-last} 
    \end{align}
    }
\end{definition}
\begin{algorithm}[t]
\caption{Causally-Fair Constrained Learning (CFCL)}
\label{algo:lagrange-training}
\DontPrintSemicolon
 \KwIn{training data $\mathcal{D}_t$, evaluation data $\mathcal{D}_e$, set $S$, precision $\epsilon$}
Set $\lambda_{\text{low}} = 0$, $\lambda_{\text{high}} =$ large \;
\While{$|\lambda_{\text{high}} - \lambda_{\text{low}}| > \epsilon$}{
    set $\lambda_{\text{mid}} = \frac{1}{2} (\lambda_{\text{low}} + \lambda_{\text{high}})$\;
    fit a neural network to solves the optimization problem in Eqs.~\ref{eq:lagrange-obj-1}-\ref{eq:lagrange-obj-last} with $\lambda = \lambda_{\text{mid}}$ on $\mathcal{D}_t$ to obtain the predictor $\widehat Y ^ S (\lambda_{\text{mid}})$ \;
    compute the causal measures of fairness NDE, NIE, NSE of $\widehat Y ^ S (\lambda_{\text{mid}})$ on evaluation data $\mathcal{D}_e$ \;
    test the hypothesis
    \begin{align} \label{eq:h0-ce}
        H_0^{CE}: \text{NCE}(\widehat y ^ S (\lambda_{\text{mid}})) &= \text{NCE}(y) \cdot \mathbb{1}(\text{CE} \notin S)
    \end{align}
    where NCE ranges in $\text{NDE}_{x_0, x_1}, \text{NIE}_{x_1, x_0}, \text{NSE}_{x_0}, \text{NSE}_{x_1}$ \\
    \textbf{if} \textit{any of $H_0^{DE}, H_0^{IE}, H_0^{SE_0}, H_0^{SE_1}$ rejected} \textbf{then} $\lambda_{\text{low}} = \lambda_{\text{mid}}$ \textbf{else}  $\lambda_{\text{high}} = \lambda_{\text{mid}}$
}
\Return{predictor $\widehat Y ^ S (\lambda_{\text{mid}})$}
\end{algorithm}
The above definition reformulates the problem of finding $\widehat Y ^ S$ to a Lagrangian form. This makes the problem amenable to standard gradient descent methods, and we propose a procedure for finding a suitable predictor $\widehat Y ^ S$ in Alg.~\ref{algo:lagrange-training} (CFCL). In principle, many non-parametric learners such as boosting or neural networks could be used for fitting $\widehat Y^S$, and here we describe a neural approach. 
The key challenge for constructing $\widehat Y ^S $ is finding the appropriate value of the tuning parameter $\lambda$. While the user may simply use a grid of values of $\lambda$, and inspect the loss function and the fairness measures, and then choose a $\lambda$ value, we propose a data-driven approach. We note that if $\lambda$ is too small, insufficient weight may be given to the fairness constraints, which therefore may be violated on new test data. If $\lambda$ is too high, however, we may give insufficient weight to minimizing the loss $L$, which may lead to poor performance on test data. Therefore, we propose a binary search type of procedure that first splits the data into train and evaluation folds, $\mathcal{D}_t$ and $\mathcal{D}_e$. CFCL starts with an interval $[\lambda_{\text{low}}, \lambda_{\text{high}}]$ and takes the midpoint $\lambda_{\text{mid}}$. For this parameter value, it computes the optimal predictor $\widehat Y ^ S (\lambda_{\text{mid}})$ for the optimization problem in Eqs.~\ref{eq:lagrange-obj-1}-\ref{eq:lagrange-obj-last} by fitting a feed-forward neural network with $n_h$ hidden layers and $n_{v}$ nodes in each layer.
Then, for this fixed value of $\lambda_{\text{mid}}$, we test the hypotheses that 
\begin{align}
    H_0^{CE}: \text{NCE}(\widehat y ^ S (\lambda_{\text{mid}})) = \text{NCE}(y) \cdot \mathbb{1}(\text{CE} \notin S)
\end{align}
on evaluation data $\mathcal{D}_e$ (done in Eq.~\ref{eq:h0-ce}), which essentially test if the fairness constrains hold on the evaluation set $\mathcal{D}_e$, i.e., out of sample.
If any of the hypotheses are rejected, it means that this value of $\lambda$ is too small to ensure that the fairness constraints are satisfied on new training data. Therefore, we want to find a larger $\lambda$, and the algorithm moves to the interval $[\lambda_{\text{mid}}, \lambda_{\text{high}}]$.
If none of the hypotheses are rejected, it means that $\lambda_{\text{mid}}$ is large enough to enforce the fairness constraints, and there may be an even smaller $\lambda$ that achieves this, so the algorithm moves to the interval $[\lambda_{\text{low}}, \lambda_{\text{mid}}]$. In this way, CFCL leads to a data-driven way to select the tuning parameter $\lambda$. As the number of training and evaluation samples increases $|\mathcal{D}_t|, |\mathcal{D}_e| \to \infty$, the method is expected to perform increasingly well. An alternative approach would be to use a framework that automatically allows the learning of the $\lambda$ parameter \citep{fioretto2021lagrangian}. 
\newcommand{\cmbvals}[2]{\footnotesize(\textcolor{blue}{#1}, \textcolor{red}{#2})}
\renewcommand{\eTOd}{\cmbvals{0}{0.2}} 
\renewcommand{\eTOi}{\cmbvals{0.06}{0.14}} 
\renewcommand{\eTOs}{\cmbvals{0.03}{-0.07}} 
\renewcommand{\dTOdi}{\cmbvals{0.07}{0.15}} 
\renewcommand{\dTOds}{\cmbvals{0.02}{-0.06}} 
\renewcommand{\iTOdi}{\cmbvals{0.02}{0.2}} 
\renewcommand{\iTOis}{\cmbvals{0.03}{-0.06}} 
\renewcommand{\diTOdis}{\cmbvals{0.02}{-0.06}} 
\renewcommand{\sTOds}{\cmbvals{-0.01}{0.2}} 
\renewcommand{\sTOis}{\cmbvals{0.06}{0.15}} 
\renewcommand{\dsTOdis}{\cmbvals{0.07}{0.15}} 
\renewcommand{\isTOdis}{\cmbvals{0}{0.2}} 
\begin{figure*}[t!]
\centering
    \begin{subfigure}[b]{0.48\textwidth}
    \centering
    \includegraphics[width=\textwidth]{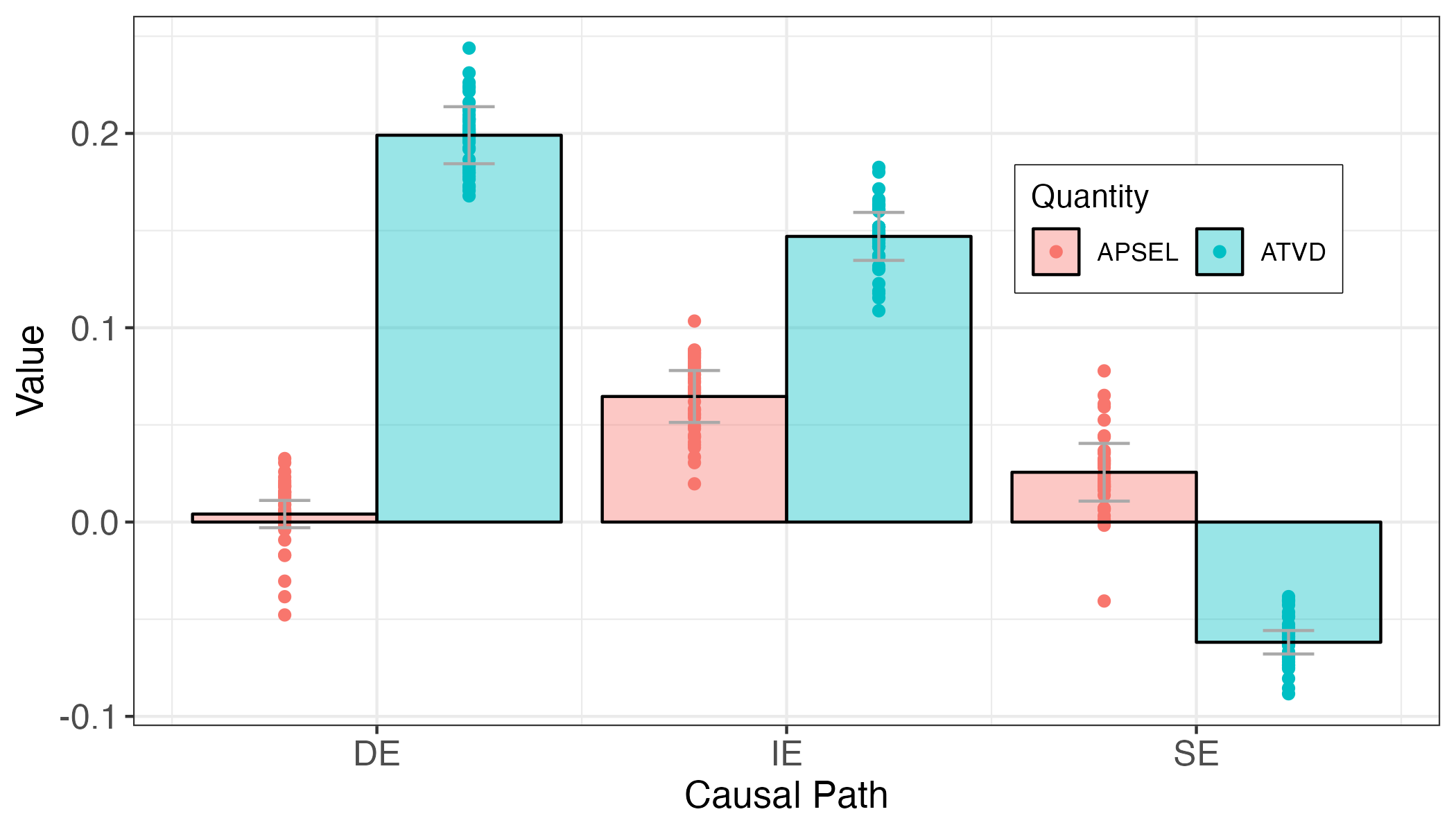}
    \caption{(A)PSEL and (A)TVD values.}
    \label{fig:apsel-census}
\end{subfigure}
\hfill
\begin{subfigure}[b]{0.48\textwidth}
    \centering
    \vspace{-0.2in}
    \includegraphics[width=\textwidth]{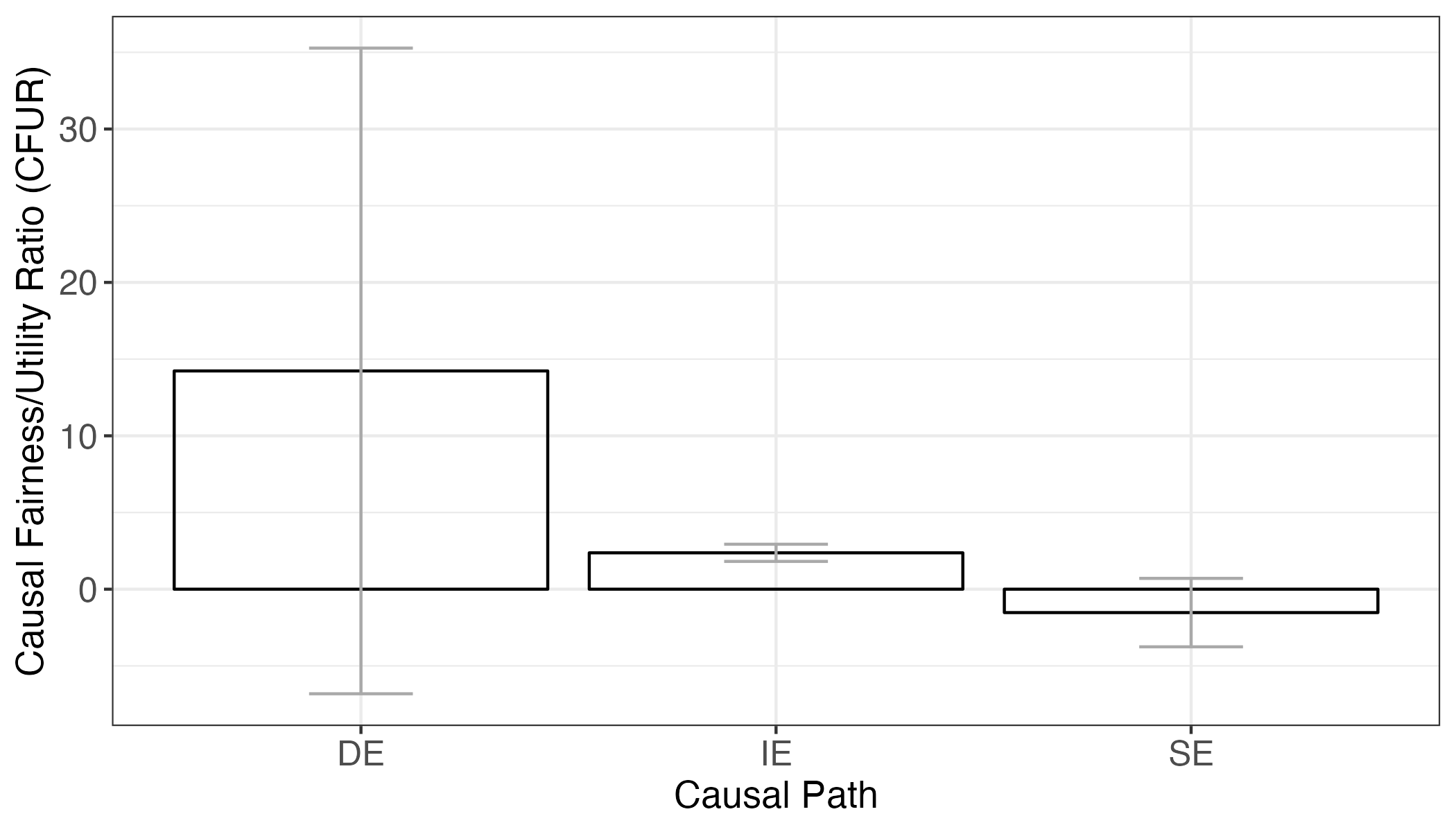}
    \caption{CFUR values.}
    \label{fig:cfur-census}
\end{subfigure}
\hfill
\begin{subfigure}[b]{0.48\textwidth}
    \centering
    \includegraphics[width=\textwidth]{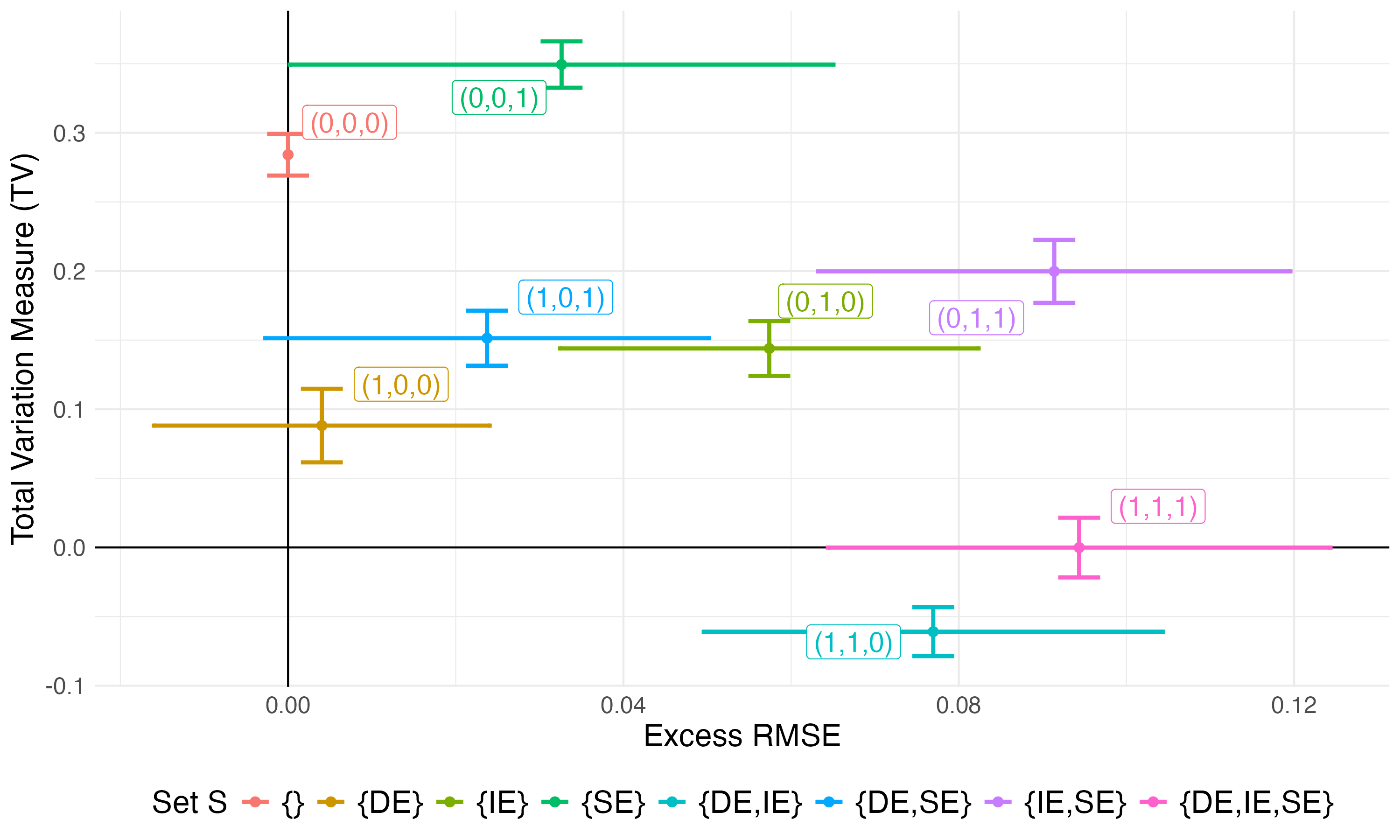} 
    \caption{Fairness/Utility of $\widehat Y^S$ predictors.}
    \label{fig:pareto-census}
\end{subfigure}
\hfill
\begin{subfigure}[b]{0.48\textwidth}
    \centering
    \scalebox{0.85}{
    \begin{tikzpicture}[node distance=2.5cm]

\node (empty) {$\emptyset$};
\node[right of=empty, xshift=0.5cm] (I) {$\{I\}$};
\node[above of=I] (D) {$\{D\}$};
\node[below of=I] (S) {$\{S\}$};
\node[right of=D] (DI) {$\{D, I\}$};
\node[right of=S] (IS) {$\{I, S\}$};
\node[right of=I] (DS) {$\{D, S\}$};
\node[right of=DS, xshift=0.5cm] (DIS) {$\{D, I, S\}$};

\draw[->] (empty) -- (D) node[sloped, above, pos=\posmid] {\eTOd};
\draw[->] (empty) -- (I) node[sloped, above, pos=\posmid] {\eTOi};
\draw[->] (empty) -- (S) node[sloped, above, pos=\posmid] {\eTOs};

\draw[->] (S) -- (IS) node[sloped, above, pos=\posmid] {\sTOis};
\draw[->] (D) -- (DI) node[sloped, above, pos=\posmid] {\dTOdi};

\draw[->] (I) -- (DI) node[sloped, above, pos=\pospart] {\iTOdi};
\draw[->] (I) -- (IS) node[sloped, above, pos=\pospart] {\iTOis};
\draw[->] (D) -- (DS) node[sloped, above, pos=\pospart] {\dTOds};
\draw[->] (S) -- (DS) node[sloped, above, pos=\pospart] {\sTOds};

\draw[->] (DI) -- (DIS) node[sloped, above, pos=\posmid] {\diTOdis};
\draw[->] (IS) -- (DIS) node[sloped, above, pos=\posmid] {\isTOdis};
\draw[->] (DS) -- (DIS) node[sloped, above, pos=\posmid] {\dsTOdis};

\end{tikzpicture}
    }
    \caption{$\gpsel$ with PSEL (blue) and TVD values (red).}
    \label{fig:gpsel-census}
\end{subfigure}
\caption{Application of Alg.~\ref{algo:pspl-attribution} on the Census 2018 dataset. (a) Estimated APSEL (Eq.~\ref{eq:apsel-compute}) and ATVD (Eq.~\ref{eq:atvd-compute}) values; (b) The causal fairness-utility ratios (Eq.~\ref{eq:cfur}); (c) The Pareto plot for trade-offs between fairness (TV measure on vertical axis) and utility (excess RMSE on horizontal axis) for different predictors. The vector ($s_1, s_2, s_3$) indicates which of the DE/IE/SE pathways are constrained to zero; (d) The $\gpsel$ graph from Fig.~\ref{fig:gpsel-general} populated with PSEL and TVD values.} 
\label{fig:census-alg-1}
\end{figure*}
\section{Experiments} \label{sec:experiments}
In this section, we perform the causal fairness-accuracy analysis described in Sec.~\ref{sec:psel} on the Census 2018 dataset (Ex.~\ref{ex:census-2018}). Additional analyses of the COMPAS (Ex.~\ref{ex:compas}) and UCI Credit (Ex.~\ref{ex:credit}) datasets are reported in Appendix~\ref{appendix:experiments}. All code for reproducing the experiments can be found in our Github repository \url{https://github.com/dplecko/causal-acc-decomp}.
\begin{example}[Salary Increase of Government Employees \citep{plecko2022causal}] \label{ex:census-2018}
    \upshape
    The US government is building a tool for automated allocation of salaries for new employees. For developing the tool, they use the data collected by the United States Census Bureau in 2018, including 
    \begin{itemize}
        \item confounders $Z$, consisting of demographic information ($Z_1$ for age, $Z_2$ for race, $Z_3$ for nationality),
        \item gender $X$ ($x_0$ female, $x_1$ male),
        \item mediators $W$, including marital and family status $M$, education $L$, and work-related information $R$,
        \item outcome $Y$, salary.
    \end{itemize}
    The government wants to predict the outcome $Y$, the yearly salary of the employees (transformed to a log-scale), in order to assign salaries for prospective employees.
    The standard fairness model (Fig.~\ref{fig:sfm}) is constructed as $\{X = X, Z = \{Z_1, Z_2, Z_3\}, W = \{M, L, R\}, Y = Y \}$.

    The team developing the ML predictor is also concerned with the fairness of the allocated salaries. In particular, they wish to understand how the different causal effects from the protected attribute $X$ to the predictor $\widehat Y$ affect the prediction, and how much the salary predictions would have to deviate from the optimal prediction to remove an effect along a specific pathway (in particular, they focus on the root mean squared error (RMSE) loss). For analyzing this, they utilize the tools from Alg.~\ref{algo:pspl-attribution}, and build causally fair predictors $\widehat Y ^S$ (for different choices of $S$-sets) using Alg.~\ref{algo:lagrange-training}.

    The analysis results are shown in Fig.~\ref{fig:census-alg-1}, with uncertainty bars indicating standard deviations over 10 bootstrap repetitions. In the analysis of PSEL and TVD values (Fig.~\ref{fig:apsel-census}), the team notices that imposing fairness constraints does not reduce RMSE for any of the effects. The largest excess loss is observed for the indirect effect, and smaller excess losses for direct and spurious effects. 
    When looking at TVD values, they find that removing direct and indirect effects reduces the group differences substantially. In terms of causal fairness-utility ratios (Fig.~\ref{fig:cfur-census}), the team finds that removing the direct effect has the best value in terms of reducing the disparity between groups vs. increasing the loss. 
    The TV measure vs. excess loss dependence for different predictors $\widehat Y^S$ is shown in Fig.~\ref{fig:pareto-census} (binary labels (D, I, S) in the figure indicate which effects were removed). The graph $\gpsel$ with the values of PSEL and TVD for each transition is shown in Fig.~\ref{fig:gpsel-census}. Based on the analysis, the team realizes that it is possible to substantially reduce discrimination with a small amount of excess RMSE loss. They decide to implement the predictor $\widehat Y ^ D$ with the direct effect removed.
\end{example}
\section{Conclusion} \label{sec:conclusion}
The tension between fairness and accuracy is a fundamental concern in the modern applications of machine learning. The importance of this tension is also recognized in the legal frameworks of anti-discrimination, such as the disparate impact doctrine, which may allow for the usage of covariates correlated with the protected attribute if they are sufficiently important for the decision-maker's utility (this concept is known as business necessity). 
In this work, we developed tools for analyzing the fairness-accuracy trade-off from a causal standpoint. Our approach allows the system designer to quantify how much excess loss is incurred when removing a path-specific causal effect (Def.~\ref{def:psel}). 
We also showed how the total excess loss, defined as the difference between the loss of the predictor fair along all causal pathways vs. an unconstrained predictor, can be decomposed into a sum of path-specific excess losses (Thm.~\ref{thm:tpl-decomposition}). 
Based on this, we developed an algorithm for attributing excess loss to different causal pathways (Alg.~\ref{algo:pspl-attribution}), and introduced the notion of a causal fairness-utility ratio that captures the $$\frac{\text{fairness gain}}{\text{excess loss}}$$ ratio and in this way summarizes the trade-off for each causal path. 
Since our approach requires access to causally-fair predictors (Def.~\ref{def:cfair-predictor}), we introduced a new neural approach for constructing such predictors (Def.~\ref{def:lagrange-form}, Alg.~\ref{algo:lagrange-training}). 
Finally, we analyzed several real-world datasets, in order to investigate if fairness and utility are in a trade-off in practice. Our findings are that, from causal perspective, fairness and utility are almost always in tension (see Exs.~\ref{ex:census-2018}-\ref{ex:credit}), contrary to some other works appearing in the fair ML literature.

\newpage
\section*{Acknowledgements} This research was supported in part by the NSF, ONR, AFOSR, DoE, Amazon, JP Morgan, and The Alfred P. Sloan Foundation.
\bibliography{refs}

\newpage
\appendix
\section*{\centering\Large Technical Appendices for \textit{\ttl}}
The source code for reproducing all the experiments can be found in our Github code repository \url{https://github.com/dplecko/causal-acc-decomp}. All experiments were performed on a MacBook Pro, with the M3 Pro chip and 36 GB RAM on macOS 14.1 (Sonoma). Each experiment can be run with less than 1 hour of compute on the above-described machine or equivalent. Within each experiment, calls to Alg.~\ref{algo:lagrange-training} are included, which includes optimizing neural networks (using \texttt{pytorch}). Here, we use feed-forward networks with $n_l = 2$ hidden layers, and $n_{\text{hidden}}=16$ nodes in each layer. Training is carried out using the Adam optimizer \citep{kingma2014adam}, with the learning rate $\eta = 0.001$ fixed, for 500 epochs with batch size 512, and early stopping regularization (20 epochs of patience), and 5 random initial restarts. 
\section{Linear Path-Specific Excess Losses} \label{appendix:linear-psel}
Consider the SCM in Eqs.~\ref{eq:lfacto-scm-1}-\ref{eq:lfacto-scm-3}. The unconstrained, DE-fair, IE-fair, and fully-fair predictors are given by
\begin{align}
    \widehat Y^\emptyset &= \alpha X + \gamma W \\
    \widehat Y^{\text{DE}} &= \gamma W \\
    \widehat Y^{\text{IE}} &= \alpha X \\
    \widehat Y^{\{ \text{DE, IE} \}} &= 0.
\end{align}
Note that $\ex [Y - \widehat Y^{\{ \text{DE, IE} \}}]^2$ equals:
\begin{align}
     &= \ex Y^2 \\
    &= \ex [\alpha X + \gamma W + \epsilon_y]^2 \\
    &= \ex[\epsilon_y^2] + \alpha^2 \ex X^2 + \gamma^2 \ex W^2 + 2 \ex \epsilon_w \ex \alpha X  \\ & \quad + \ex \epsilon_w \ex \gamma W \quad + 2 \ex \alpha X \ex \gamma W \\
    &= \sigma_y^2 + \frac{\alpha^2}{2} + \gamma^2 \ex [\beta X + \epsilon_w]^2 + 2 \frac{\alpha}{2} \ex \gamma[\beta X + \epsilon_w] \\
    &=  \sigma_y^2 + \frac{\alpha^2}{2} + \gamma^2 \sigma_w^2 + \frac{\gamma^2\beta^2}{2} + \frac{\alpha\gamma\beta}{2} \\
    &= \sigma_y^2 + \frac{\alpha^2 + \gamma^2 \beta^2 + \alpha\gamma\beta}{2} + \gamma^2 \sigma_w^2.
\end{align}
Similarly, we have that
\begin{align}
     \ex [Y - \widehat Y^{\text{DE}}]^2 &= \ex [\alpha X + \gamma W + \epsilon_y - \gamma W ]^2 \\
    &= \sigma_y^2 + \alpha^2 \ex X^2 \\
    &= \sigma_y^2 + \frac{\alpha^2}{2} \\
    \ex [Y - \widehat Y^{\text{IE}}]^2 &= \ex [\alpha X + \gamma W + \epsilon_y - \alpha X ]^2 \\
    &= \sigma_y^2 + \gamma^2 \ex W^2 \\
    &= \sigma_y^2 + \frac{\gamma^2\beta^2}{2} + \gamma^2 \sigma_w^2\\
    \ex[Y - \widehat Y^\emptyset]^2 &= \ex [\alpha X + \gamma W + \epsilon_y - \alpha X - \gamma W]^2 \\
    &= \ex \epsilon_y^2 = \sigma_y^2.
\end{align}
Therefore, we have that $\ex [Y - \widehat Y^{\{ \text{DE, IE} \}}]^2 - \ex[Y - \widehat Y^\emptyset]^2$ equals
\begin{align}
     & \frac{\alpha^2 + \gamma^2 \beta^2 + \alpha\gamma\beta}{2} + \gamma^2 \sigma_w^2 \\
    &= \frac{\alpha^2}{2} + \frac{\gamma^2 \beta^2 + \alpha\gamma\beta}{2} + \gamma^2 \sigma_w^2 \\
    &= \ex[Y - \widehat Y^\text{DE}]^2 - \ex[Y - \widehat Y^\emptyset]^2 \\
    &\quad + \ex [Y - \widehat Y^{\{ \text{DE, IE} \}}]^2 - \ex[Y - \widehat Y^\text{DE}]^2. 
\end{align}

\section{Theorem Proofs} \label{appendix:theorem-proofs}

\begin{proof}[Thm.~\ref{thm:tpl-decomposition} Proof:] The total excess loss (TEL) is defined as:
\begin{align}
    \text{PSEL}(\emptyset \to \{D, I, S\}) = \ex[L(\widehat Y^{\{D, I, S\}}, Y)] - \ex[L(\widehat Y^\emptyset, Y)].
\end{align}
Let $S_1, S_2 \subset \{D, I, S\}$ such that $S_1 \neq S_2$. Note that the quantity $\text{PSEL}(\emptyset \to \{D, I, S\})$ can be expanded as follows:
\begin{align}
    &\ex[L(\widehat Y^{\{D, I, S\}}, Y)] - \ex[L(\widehat Y^\emptyset, Y)] \\
    &= \ex[L(\widehat Y^{\{D, I, S\}}, Y)] - \ex[L(\widehat Y^{S_1}, Y)] \\
    &\quad + \ex[L(\widehat Y^{S_1}, Y)] - \ex[L(\widehat Y^\emptyset, Y)] \\
    &= \underbrace{\ex[L(\widehat Y^{\{D, I, S\}}, Y)] - \ex[L(\widehat Y^{\{S_1, S_2\}}, Y)]}_{\text{PSEL}(\{S_1, S_2\} \to \{D, I, S\})} \\
    &\quad + \underbrace{\ex[L(\widehat Y^{\{S_1, S_2\}}, Y)] - \ex[L(\widehat Y^{S_1}, Y)]}_{\text{PSEL}(S_1 \to \{S_1, S_2\})} \\
    &\quad + \underbrace{\ex[L(\widehat Y^{S_1}, Y)] - \ex[L(\widehat Y^\emptyset, Y)]}_{\text{PSEL}(\emptyset \to S_1)} \\
    &= \text{PSEL}(\emptyset \to S_1) + \text{PSEL}(S_1 \to \{S_1, S_2\}) \\ &\quad + \text{PSEL}(\{S_1, S_2\} \to \{D, I, S\}),
\end{align}
completing the theorem's proof for the choice $S_1 = D, S_2 = I$. The proof also implies other decompositions of the total excess loss $\text{PSEL}(\emptyset \to \{D, I, S\})$ as mentioned in Rem.~\ref{remark:non-unique-decomp}.
\end{proof}

\begin{proof}[Prop.~\ref{prop:pspl-shapley} Proof:] 
    Let $\mathcal{S} = S_1, \dots, S_m$ be a set of causal pathways between the attribute $X$ and the outcome $Y$ (such as direct, indirect, and spurious effects in the context of the Standard Fairness Model). Consider the value function 
    \begin{align}
        f(S) = \text{PSEL}(\emptyset \to S).
    \end{align}
    The Shapley value of pathway $S_i$ associated with $f$ is given by:
    \begin{align}
        \phi (S_i) = \sum_{S \subseteq \mathcal{S} \setminus \{S_i\}} \frac{1}{m\binom{m-1}{|S|}} \left(f(S \cup \{S_i\}) - f(S)\right).
    \end{align}
    Consider now the generalized $\gpsel$ graph for $m$ pathways shown in Fig.~\ref{fig:gpsel-general}. The graph is split into columns, and in column $\ell$ all of the sets $S$ of size $\ell$ appear. The quantity $\text{APSEL}(S_i)$ considers all paths $\pi$ that start in $\emptyset$ and end in $\{S_1, \dots, S_m\}$, and takes the average of excess losses associated with $S_i$ along each path $\pi$, written $\text{PSEL}(\pi^{<S_i} \to \pi^{\leq S_i})$. The APSEL quantity is written as:
    \begin{align}
        \frac{1}{n!} \sum_{\substack{\text{paths } \pi \in \gpsel: \\ \emptyset \text{ to } \{D, I, S\}}} \text{PSEL}(\pi^{<S_i}  \to \pi^{\leq S_i}).
    \end{align}
    We now split the expression for the Shapley value $\phi (S_i)$ by conditioning on the size of the set $S$:
    \begin{align}
         & \sum_{\ell = 0}^{m-1} \sum_{S \subseteq \mathcal{S} \setminus \{S_i\}: |S|=\ell} \frac{1}{m\binom{m-1}{\ell}} \left(f(S \cup \{S_i\}) - f(S)\right) \\
        &= \sum_{\ell = 0}^{m-1} \sum_{S \subseteq \mathcal{S} \setminus \{S_i\}: |S|=\ell} \frac{1}{m\binom{m-1}{\ell}} \big(\text{PSEL}(\emptyset \to \{S \cup S_i\}) \\
        & \qquad \qquad \qquad \qquad \qquad \qquad \qquad- \text{PSEL}(\emptyset \to S) \big) \nonumber \\
        &= \sum_{\ell = 0}^{m-1} \sum_{S \subseteq \mathcal{S} \setminus \{S_i\}: |S|=\ell} \frac{1}{m\binom{m-1}{\ell}} \text{PSEL}(S \to \{S \cup S_i\}).
    \end{align}
    Now, fix a set $S$ of size $|S| = \ell$. There are exactly $\ell!$ paths leading from the $\emptyset$ to $S$. This is followed by a fixed transition $S \to S \cup S_i$ is chosen. The remaining steps from $S \cup S_i$ to $S_1, \dots, S_m$ along the path $\pi$ can be chosen in $(m-\ell-1)!$ ways. Therefore, across all paths $\pi$, the contribution $\text{PSEL}(S \to \{S \cup S_i\}$ appears $\ell!(m-\ell-1)!$ times. Therefore, the APSEL$(S_i)$ quantity can be re-written as:
    \begin{align}
        & \frac{1}{m!} \sum_{\ell = 0}^{m-1} \sum_{S \subseteq \mathcal{S} \setminus \{S_i\}: |S|=\ell} {\ell!(n-\ell -1)!} \text{PSEL}(S \to \{S \cup S_i\}) \\
        &=  \sum_{\ell = 0}^{m-1} \sum_{S \subseteq \mathcal{S} \setminus \{S_i\}: |S|=\ell} \frac{\ell!(m-\ell -1)!}{m!} \text{PSEL}(S \to \{S \cup S_i\}) \\
        &= \sum_{\ell = 0}^{m-1} \sum_{S \subseteq \mathcal{S} \setminus \{S_i\}: |S|=\ell} \frac{1}{m \binom{m-1}{\ell}} \text{PSEL}(S \to \{S \cup S_i\}) \\
        &= \sum_{S \subseteq \mathcal{S} \setminus \{S_i\}} \frac{1}{m \binom{m-1}{|S|}} \text{PSEL}(S \to \{S \cup S_i\}) \\
        &= \phi(S_i).
    \end{align}
The proof for the ATVD quantity follows the same steps.
\end{proof}
\begin{figure}[t]
    \centering
    \scalebox{0.88}{
    \begin{tikzpicture}[node distance=1.5cm]

\node (empty) {$\emptyset$};
\node[right of=empty, xshift=0.5cm] (I) {$\vdots$};
\node[above of=I] (D) {$\{S_1\}$};
\node[below of=I] (S) {$\{S_m\}$};
\node[right of=D, xshift = 1.25cm] (DI) {$\{S_1, \dots, S_{\ell}\}$};
\node[right of=S, xshift = 1.25cm] (IS) {$\{S_{m-\ell+1}, \dots,  S_m\}$};
\node[right of=I, xshift = 1.25cm] (DS) {$\vdots$};
\node[right of=DS, xshift=1.25cm] (DIS) {$\{S_1, \dots, S_m\}$};

\node[draw, fill=gray!30, below of=empty, yshift=-1cm] (col1) {Column 0};
\node[draw, fill=gray!30, below of=S, yshift=0.5cm] (col2) {Column 1};
\node[draw, fill=gray!30, below of=IS, yshift=0.5cm] (coll) {Column $\ell$};
\node[draw, fill=gray!30, below of=DIS, yshift=-1cm] (colm) {Column $m$};

\path (col2) -- (coll) node[midway] {$\dots$};
\path (coll) -- (colm) node[midway] {$\dots$};

\draw[->] (empty) -- (D) node[sloped, above, pos=\posmid] {};
\draw[->] (empty) -- (I) node[sloped, above, pos=\posmid] {};
\draw[->] (empty) -- (S) node[sloped, above, pos=\posmid] {};

\draw[->] (S) -- (IS) node[sloped, above, pos=\posmid] {$\dots$};
\draw[->] (D) -- (DI) node[sloped, above, pos=\posmid] {$\dots$};

\draw[->] (I) -- (DI) node[sloped, above, pos=\pospart] {$\dots$};
\draw[->] (I) -- (IS) node[sloped, above, pos=\pospart] {$\dots$};
\draw[->] (D) -- (DS) node[sloped, above, pos=\pospart] {$\dots$};
\draw[->] (S) -- (DS) node[sloped, above, pos=\pospart] {$\dots$};

\draw[->] (DI) -- (DIS) node[sloped, above, pos=\posmid] {$\dots$};
\draw[->] (IS) -- (DIS) node[sloped, above, pos=\posmid] {$\dots$};
\draw[->] (DS) -- (DIS) node[sloped, above, pos=\posmid] {$\dots$};

\end{tikzpicture}
    }
    \caption{Generalized $\gpsel$ graph for $m$ causal pathways.}
    \label{fig:gpsel-general}
\end{figure}
\section{Connection to Causal Decompositions} \label{appendix:causal-decomposition}
\begin{figure*}
    \begin{subfigure}{0.48\textwidth}
        \centering
        \includegraphics[width=\linewidth]{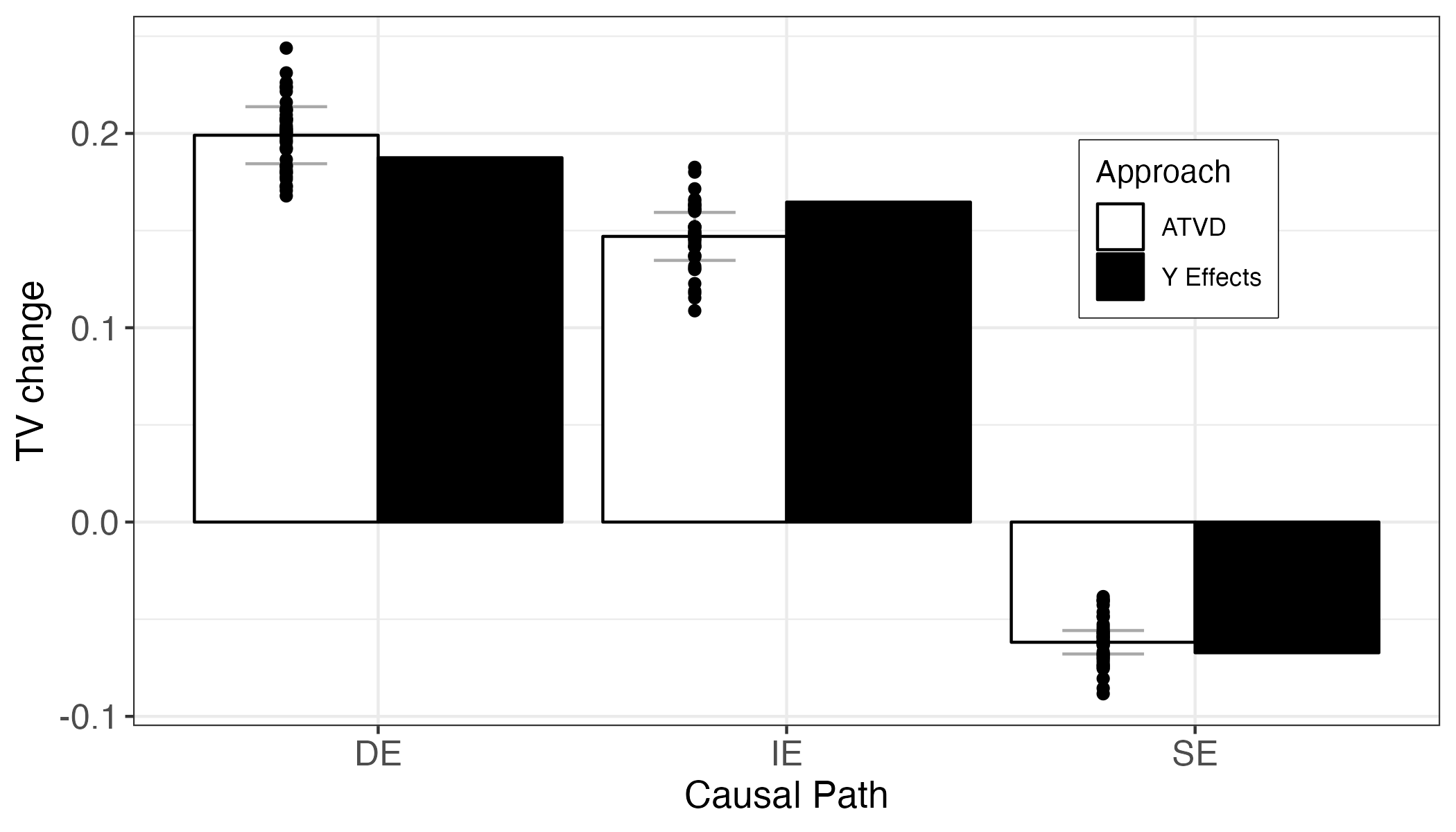}
        \caption{TVD and TVR comparison on Census 2018 data.}
        \label{fig:tv-bar-census}   
    \end{subfigure}
    \hfill
    \begin{subfigure}{0.48\textwidth}
        \centering
        \includegraphics[width=\linewidth]{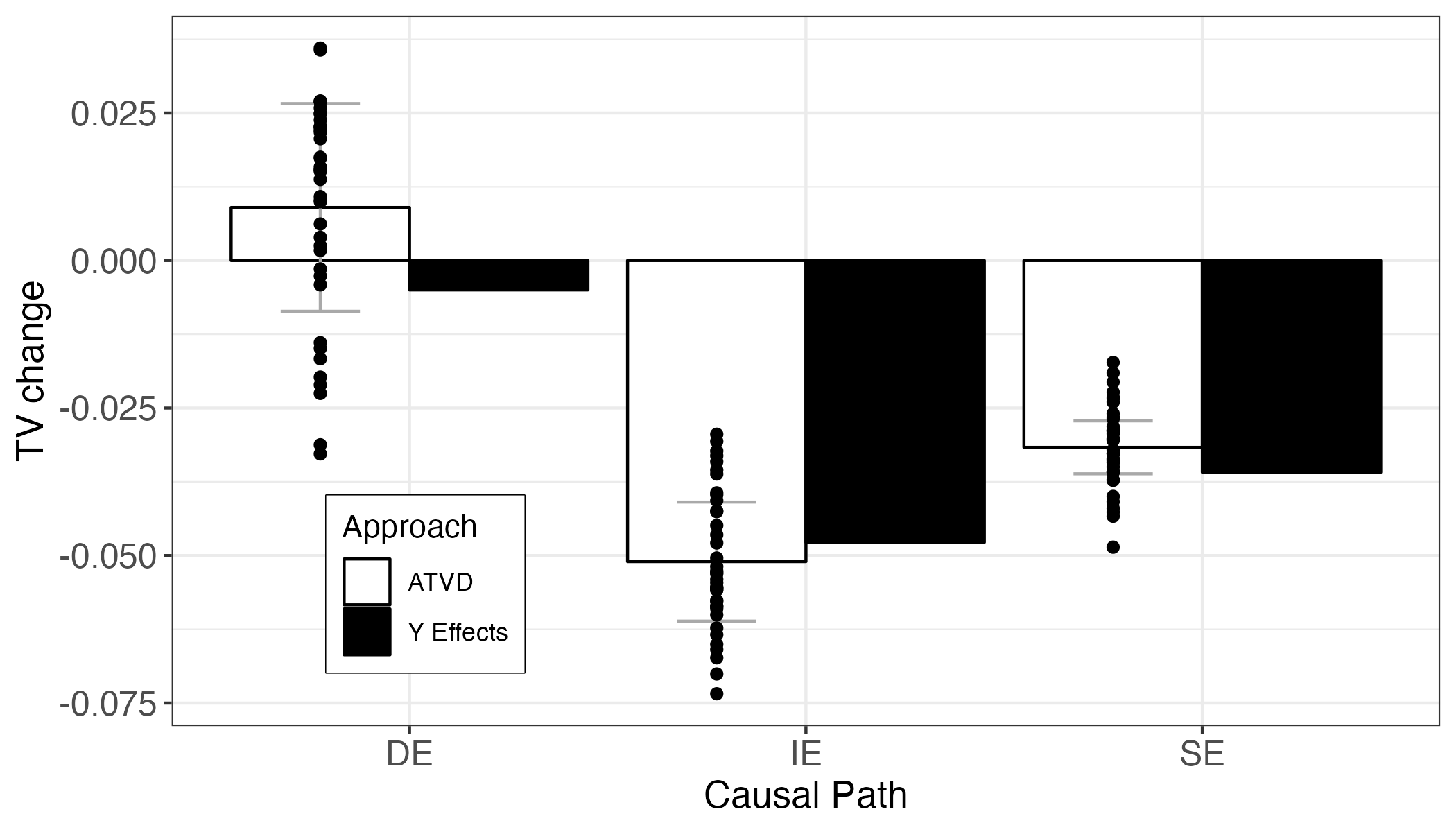}
        \caption{TVD and TVR comparison on COMPAS data.}
        \label{fig:tv-bar-compas}
    \end{subfigure}
    \caption{Comparison of alternative causal disparity quantifications TVD, TVR.}
    \label{fig:tv-bars}
\end{figure*}
In this appendix, we explore the connection of the fairness/accuracy trade-off described in the main text with the literature on causal decompositions \citep{zhang2018fairness, plecko2022causal}.
As described in the main text and Alg.~\ref{algo:pspl-attribution}, along with tracking the reduction in predictive power from imposing a causal fairness constraint, we were also interested in the change in the TV$_{x_0, x_1}(\widehat y^S)$, captured by the TV difference (TVD) quantity defined as:
\begin{align}
    \text{TVD}(S \to S') =& \underbrace{\ex[Y^{S'} \mid x_1] - \ex[Y^{S'} \mid x_0]}_{\text{TV after removing } S'\setminus S} \\ &- \underbrace{\ex[Y^{S} \mid x_1] - \ex[Y^{S} \mid x_0]}_{\text{TV before removing } S'\setminus S}.
\end{align}
In words, the quantity $\text{TVD}(S \to S')$ measures how much the TV measure changes once the effect in $S' \setminus S$ is constrained to $0$.
Put differently, the $\text{TVD}(S \to S')$ is a quantification of the amount of disparity that is transmitted along the path $S' \setminus S$. 
Other notions for quantifying the size of the disparity transmitted along a specific causal pathway have also been explored in the literature on causal decompositions \citep{zhang2018fairness, plecko2022causal}. An important result from \citep{plecko2022causal} we draw a connection to is the decomposition of the TV measure into the NDE, NIE, and NSE contributions:
\begin{proposition}[TV Decomposition \citep{plecko2022causal}]
    The TV measure can be decomposed into the natural direct, indirect, and spurious contributions:
    \begin{align} \label{eq:tv-decomposition}
    \text{TV}_{x_0, x_1}(y) &= \text{NDE}_{x_0, x_1}(y) - \text{NIE}_{x_1, x_0}(y) \\ &\quad + \text{NSE}_{x_1}(y) - \text{NSE}_{x_0}(y).
\end{align}
\end{proposition}
Based on the decomposition in Eq.~\ref{eq:tv-decomposition}, we can try to quantify how much the disparity would be reduced by if we were to remove an effect. For instance, if the direct effect was removed, the reduction in the disparity would be
\begin{align}
    \text{TV}_{x_0, x_1}(y^{\text{DE}}) - \text{TV}_{x_0, x_1}(y) = -\text{NDE}_{x_0, x_1}(y)
\end{align}
where $y^{\text{DE}}$ is the outcome with the direct effect removed, which we assume satisfies:
\begin{align}
    \text{NDE}_{x_0, x_1}(y^{\text{DE}}) &= 0 \\
    \text{NIE}_{x_1, x_0}(y^{\text{DE}}) &= \text{NIE}_{x_1, x_0}(y) \\
    \text{NSE}_{x_0}(y^{\text{DE}}) &= \text{NSE}_{x_0}(y) \\
    \text{NSE}_{x_1}(y^{\text{DE}}) &= \text{NSE}_{x_1}(y). \\
\end{align}
A similar reasoning could be applied to understand the reduction in the TV resulting from removing indirect, and spurious effects. Therefore, in this way we can obtain a different quantification of the TVD measure, which we call TV reduction (TVR), satisfying:
\begin{align}
    \text{TVR}(S \to S \cup \text{DE}) &= -\text{NDE}_{x_0, x_1}(y) \\
    \text{TVR}(S \to S \cup \text{IE}) &= \text{NIE}_{x_1, x_0}(y) \\
    \text{TVR}(S \to S \cup \text{SE}) &= - \text{NSE}_{x_1}(y) + \text{NSE}_{x_0}(y),
\end{align}
where in each line we assume the set $S$ does not contain DE, IE, and SE, respectively. To establish a connection between our approach with the literature on causal decompositions, we empirically compare the TVR and TVD measures. We again use the same datasets as in the main text, namely the Census 2018 dataset from Ex.~\ref{ex:census-2018} and the COMPAS dataset from Ex.~\ref{ex:compas}. 

The results of the empirical comparisons of TVR and TVD values are shown in Fig.~\ref{fig:tv-bars}, where the indicate bars are standard deviations obtained from 10 bootstrap samples of the data. The figure illustrates that on both COMPAS and Census data, TVD and TVR values are not statistically different, across direct, indirect, and spurious effects. This finding adds further validity to our analysis in the main text, meaning that the TVD measure corresponds closely with existing notions for quantifying discrimination in the literature.

\section{CFUR Quantity Representation} \label{appendix:cfur-weighted-average}
The CFUR quantity (Def.~\ref{def:cfur}) of a causal path $S_i$ is given by
\begin{align}
    \frac{\text{ATVD}(S_i)}{\text{APSEL}(S_i)}.
\end{align}
The ATVD and APSEL quantities are defined as the average values of TVD, PSEL quantities, defined through Eqs.~\ref{eq:apsel-compute}-\ref{eq:atvd-compute}. In this appendix, we discuss the connection of the CFUR value with the local fairness-utility ratios, defined via:
\begin{align}
    \text{LCFUR}(S, S_i) = \frac{\text{TVD}(S \to S \cup S_i)}{\text{PSEL}(S \to S \cup S_i)}. 
\end{align}
From Eqs.~\ref{eq:apsel-compute}-\ref{eq:atvd-compute}, we see that
\begin{align}
    \text{CFUR}(S_i) = \frac{\frac{1}{m!}\sum_{S \in \mathcal{I}} \text{TVD}(S \to S \cup S_i)}{\frac{1}{m!}\sum_{S \in \mathcal{I}} \text{PSEL}(S \to S \cup S_i)},
\end{align}
where $m$ is the number of causal paths and in the summations the set $S$ ranges within $\mathcal{I}$, the set of all path prefixes of $S_i$ along all paths connecting $\emptyset$ and $\{S_1, \dots, S_m\}$ in the $\gpsel$ graph.
Now, we want to connect the global CFUR with the local LCFUR values. This can be done through the following proposition:
\begin{proposition}[LCFUR and CFUR]
    The CFUR$(S_i)$ value of a causal path $S_i$ can be written as:
    \begin{align}
        \text{CFUR}(S_i) = \sum_{S \in \mathcal{I}} \lambda (S_i) \times \text{LCFUR}(S, S_i),
    \end{align}
    where the weights $\lambda (S, S_i)$ are defined as:
    \begin{align}
        \lambda(S, S_i) = \frac{\text{PSEL}(S \to S \cup S_i)}{\sum_{S' \in \mathcal{I}} \text{PSEL}(S' \to S' \cup S_i)}.
    \end{align}
\end{proposition}
The above proposition shows that the global CFUR value is a weighted average of all the local LCFUR values. The weight assigned to each local LCFUR is proportional to the size of its path-specific excess loss $\text{PSEL}(S \to S \cup S_i)$ among the sum of all path-specific excess losses that are considered $\sum_{S' \in \mathcal{I}} \text{PSEL}(S' \to S' \cup S_i)$. Therefore, we see that the CFUR value places larger weights on $S \to S \cup S_i$ transitions that result in a larger path-specific excess loss. 
\section{Additional Experiments} \label{appendix:experiments}
In this appendix, we analyze the COMPAS dataset \citep{larson2016how} (Ex.~\ref{ex:compas}), and the UCI Credit dataset \citep{uci2016credit} (Ex.~\ref{ex:credit}). 
\renewcommand{\eTOd}{\cmbvals{0}{0.01}} 
\renewcommand{\eTOi}{\cmbvals{0.1}{-0.05}} 
\renewcommand{\eTOs}{\cmbvals{0.04}{-0.03}} 
\renewcommand{\dTOdi}{\cmbvals{0.09}{-0.05}} 
\renewcommand{\dTOds}{\cmbvals{0.03}{-0.03}} 
\renewcommand{\iTOdi}{\cmbvals{-0.01}{0.01}} 
\renewcommand{\iTOis}{\cmbvals{0.09}{-0.03}} 
\renewcommand{\diTOdis}{\cmbvals{0.06}{-0.03}} 
\renewcommand{\sTOds}{\cmbvals{0}{0.01}} 
\renewcommand{\sTOis}{\cmbvals{0.15}{-0.05}} 
\renewcommand{\dsTOdis}{\cmbvals{0.12}{-0.05}} 
\renewcommand{\isTOdis}{\cmbvals{-0.04}{0.01}} 
\begin{figure*}[t]
\centering
\begin{subfigure}[b]{0.49\textwidth}
    \centering
    \includegraphics[width=\textwidth]{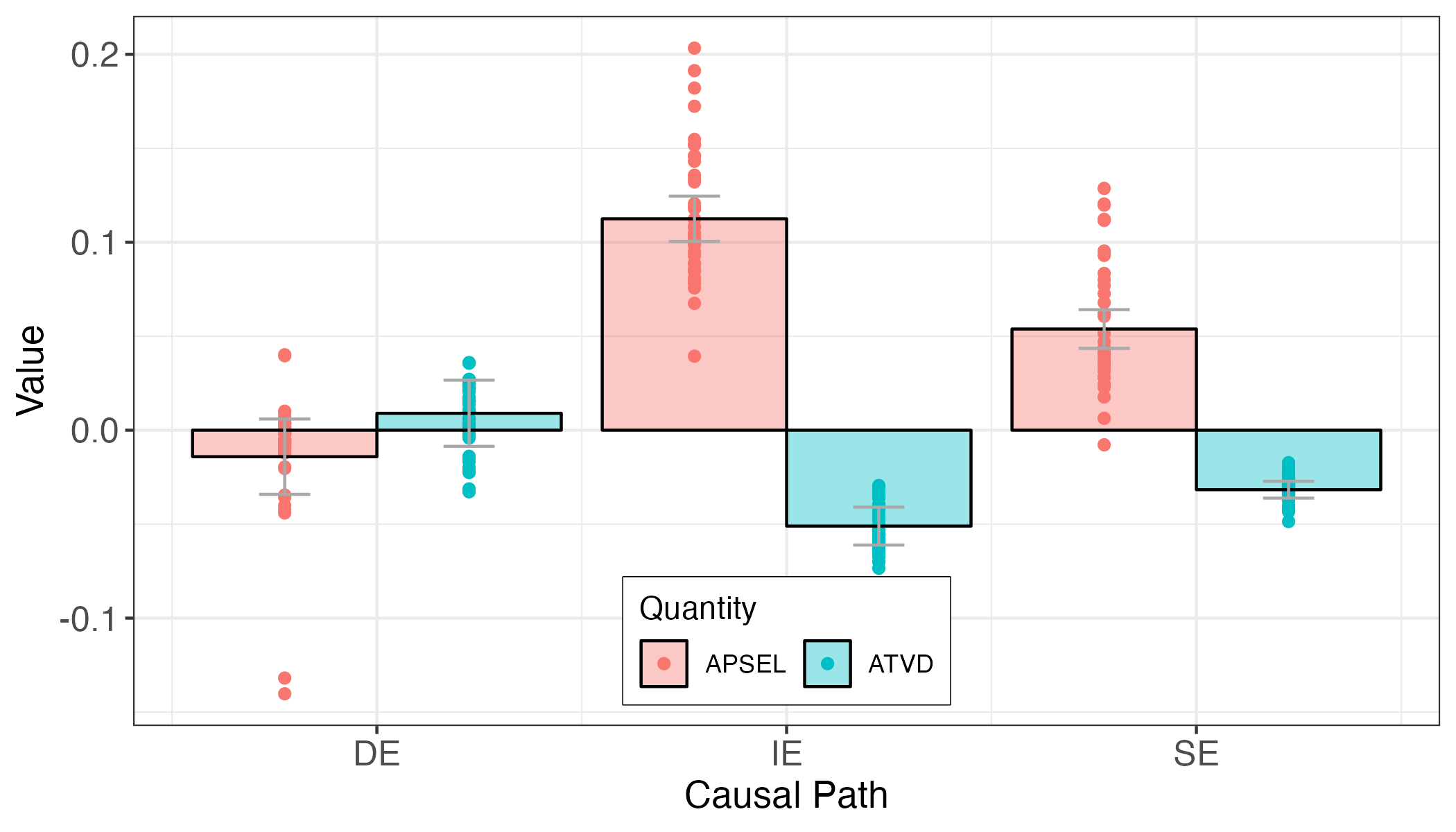}
    \caption{(A)PSEL and (A)TVD values.}
    \label{fig:apsel-compas}
\end{subfigure}
\hfill
\begin{subfigure}[b]{0.49\textwidth}
    \centering
    \vspace{-0.2in}
    \includegraphics[width=\textwidth]{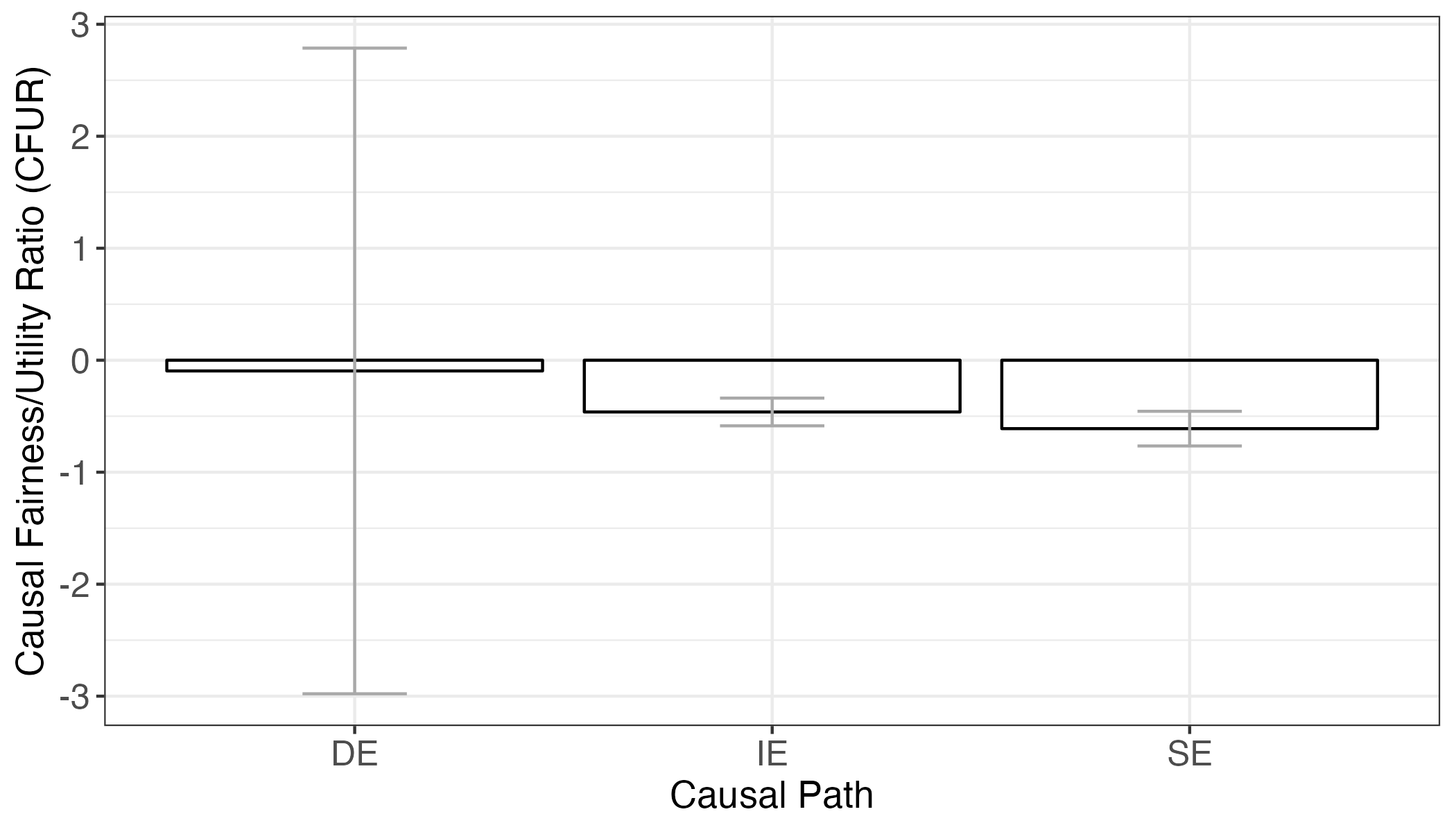}
    \caption{CFUR values.}
    \label{fig:cfur-compas}
\end{subfigure}
\newline
\begin{subfigure}[b]{0.54\textwidth}
    \centering
    \includegraphics[width=\textwidth]{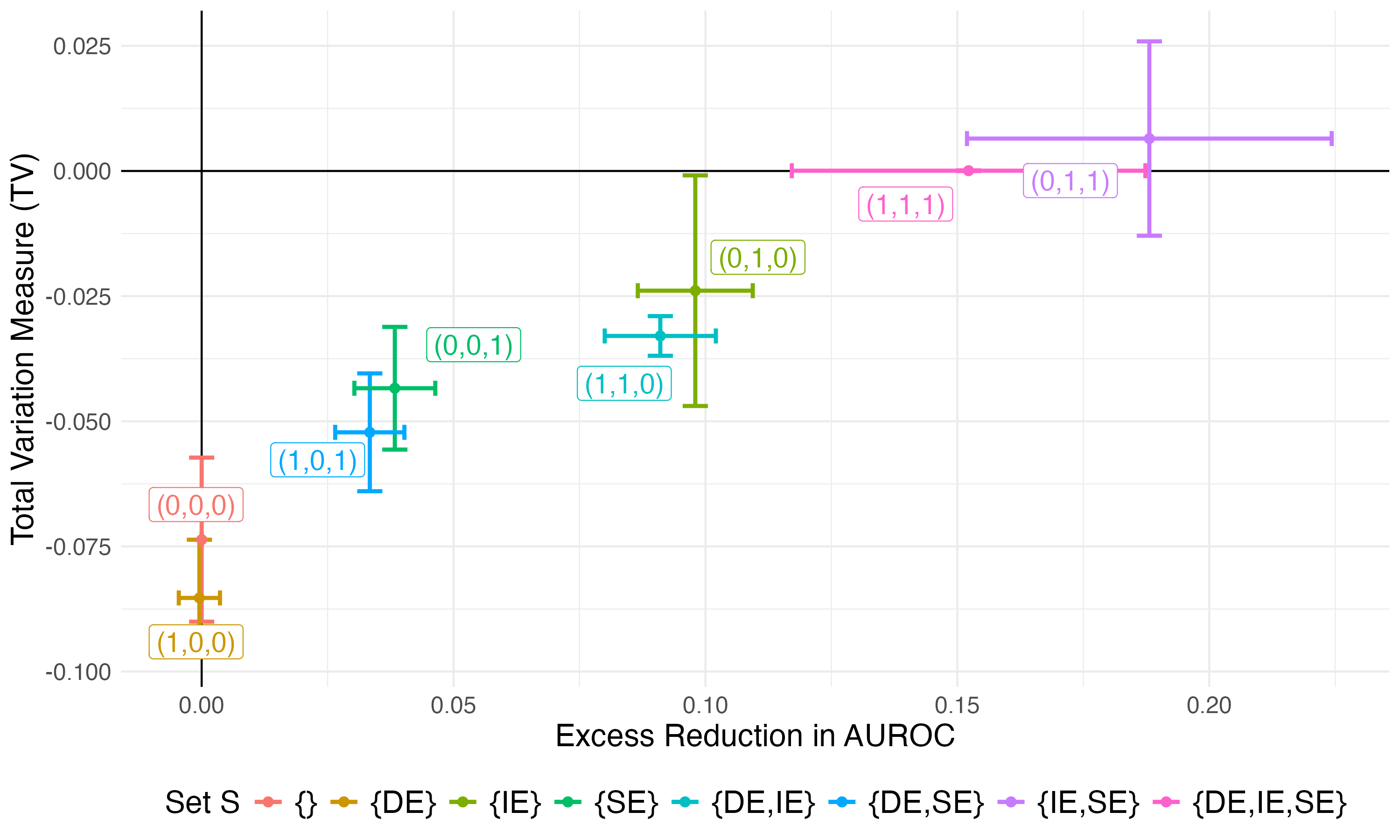}
    \caption{Fairness/Utility of $\widehat Y^S$ predictors.}
    \label{fig:pareto-compas}
\end{subfigure}
\hfill
\begin{subfigure}[b]{0.45\textwidth}
    \centering
    \scalebox{0.85}{
    \begin{tikzpicture}[node distance=2.5cm]

\node (empty) {$\emptyset$};
\node[right of=empty, xshift=0.5cm] (I) {$\{I\}$};
\node[above of=I] (D) {$\{D\}$};
\node[below of=I] (S) {$\{S\}$};
\node[right of=D] (DI) {$\{D, I\}$};
\node[right of=S] (IS) {$\{I, S\}$};
\node[right of=I] (DS) {$\{D, S\}$};
\node[right of=DS, xshift=0.5cm] (DIS) {$\{D, I, S\}$};

\draw[->] (empty) -- (D) node[sloped, above, pos=\posmid] {\eTOd};
\draw[->] (empty) -- (I) node[sloped, above, pos=\posmid] {\eTOi};
\draw[->] (empty) -- (S) node[sloped, above, pos=\posmid] {\eTOs};

\draw[->] (S) -- (IS) node[sloped, above, pos=\posmid] {\sTOis};
\draw[->] (D) -- (DI) node[sloped, above, pos=\posmid] {\dTOdi};

\draw[->] (I) -- (DI) node[sloped, above, pos=\pospart] {\iTOdi};
\draw[->] (I) -- (IS) node[sloped, above, pos=\pospart] {\iTOis};
\draw[->] (D) -- (DS) node[sloped, above, pos=\pospart] {\dTOds};
\draw[->] (S) -- (DS) node[sloped, above, pos=\pospart] {\sTOds};

\draw[->] (DI) -- (DIS) node[sloped, above, pos=\posmid] {\diTOdis};
\draw[->] (IS) -- (DIS) node[sloped, above, pos=\posmid] {\isTOdis};
\draw[->] (DS) -- (DIS) node[sloped, above, pos=\posmid] {\dsTOdis};

\end{tikzpicture}
    }
    \caption{$\gpsel$ with PSEL (blue) and TVD values (red).}
    \label{fig:gpsel-compas}
\end{subfigure}
\caption{Application of Alg.~\ref{algo:pspl-attribution} on the COMPAS dataset.}
\label{fig:compas-alg-1}
\end{figure*}
\begin{example}[Recidivism Prediction on COMPAS \citep{larson2016how}] \label{ex:compas}
    \upshape
    Courts in Broward County, Florida use machine learning algorithms, developed by a private company, to predict whether individuals released on parole are at high risk of re-offending within 2 years ($Y$). The algorithm is based on the demographic information $Z$ ($Z_1$ for gender, $Z_2$ for age), race $X$ ($x_0$ denoting White, $x_1$ Non-White), juvenile offense counts $J$, prior offense count $P$, and degree of charge $D$. The grouping $\{X = X, Z = \{Z_1, Z_2\}, W = \{J, P, D\}, Y =  Y \}$ constructs the standard fairness model.

    The team from ProPublica wishes to understand the tension between fairness and accuracy, since the latter is an important concern of the district court. For the analysis, they use the complement of the area under the receiver operator characteristic curve ($1-$AUROC). They apply Alg.~\ref{algo:pspl-attribution}, and find that the APSEL value is the largest for the indirect effect (around 12\% of AUROC), while it is smaller for the spurious effect (6\%), and negligible for the direct effect, implying that the direct effect does not play a significant role in the prediction (see Fig.~\ref{fig:apsel-compas}). Regarding TVD values, the team finds that the removal of the spurious effect results in a 3\% decrease in the TV measure, while the removal of the indirect effect decreases the disparity by 5\%. Based on this, they find that the spurious effect is the best for reducing the disparity between groups (see Fig.~\ref{fig:cfur-compas} for CFUR values). 
    Finally, the team also visualizes the TV and excess loss dependence (Fig.~\ref{fig:pareto-compas}), and the graph $\gpsel$ with PSEL and TVD values associated with each transition and effect removal (Fig.~\ref{fig:apsel-compas}). In the upcoming court hearing, the team proposes the usage of the causally-constrained predictor $\widehat Y ^ {\{D, S\}}$ that removes direct and spurious effects, and they use this analysis to demonstrate the impact of this choice on accuracy.
\end{example}

\renewcommand{\eTOd}{\cmbvals{0.01}{0}} 
\renewcommand{\eTOi}{\cmbvals{0.05}{0.01}} 
\renewcommand{\eTOs}{\cmbvals{0}{0}} 
\renewcommand{\dTOdi}{\cmbvals{0.04}{0.01}} 
\renewcommand{\dTOds}{\cmbvals{0}{0}} 
\renewcommand{\iTOdi}{\cmbvals{0.01}{0}} 
\renewcommand{\iTOis}{\cmbvals{0}{0}} 
\renewcommand{\diTOdis}{\cmbvals{0}{0}} 
\renewcommand{\sTOds}{\cmbvals{0}{0}} 
\renewcommand{\sTOis}{\cmbvals{0.04}{0.01}} 
\renewcommand{\dsTOdis}{\cmbvals{0.05}{0.01}} 
\renewcommand{\isTOdis}{\cmbvals{0.01}{0}} 
\begin{figure*}[t]
\centering
    \begin{subfigure}[b]{0.49\textwidth}
    \centering
    \includegraphics[width=\textwidth]{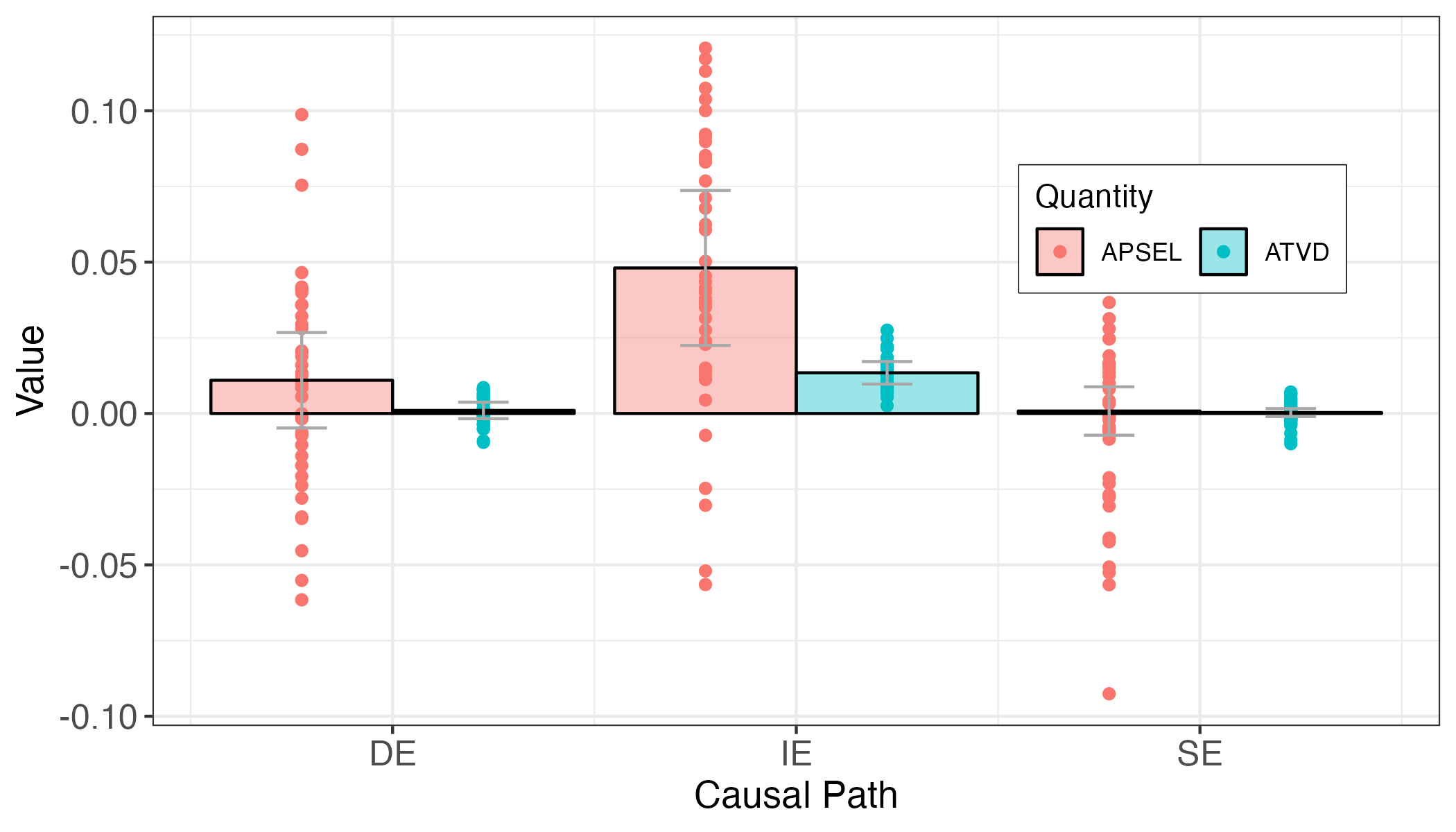} 
    \caption{(A)PSEL and (A)TVD values.}
    \label{fig:apsel-credit}
\end{subfigure}
\hfill
\begin{subfigure}[b]{0.49\textwidth}
    \centering
    \vspace{-0.2in}
    \includegraphics[width=\textwidth]{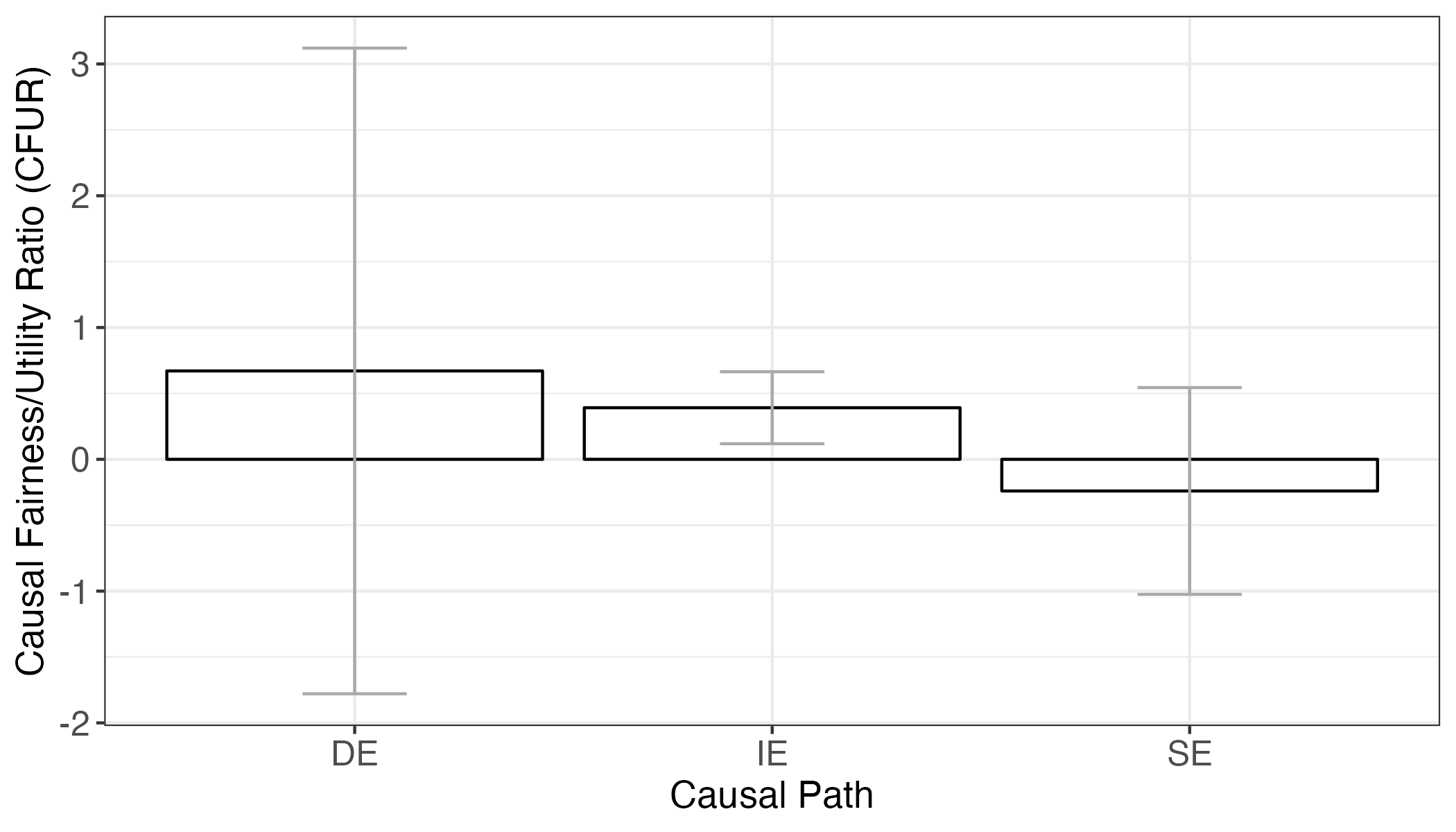} 
    \caption{CFUR values.}
    \label{fig:cfur-credit}
\end{subfigure}
\newline
\begin{subfigure}[b]{0.54\textwidth}
    \centering
    \includegraphics[width=\textwidth]{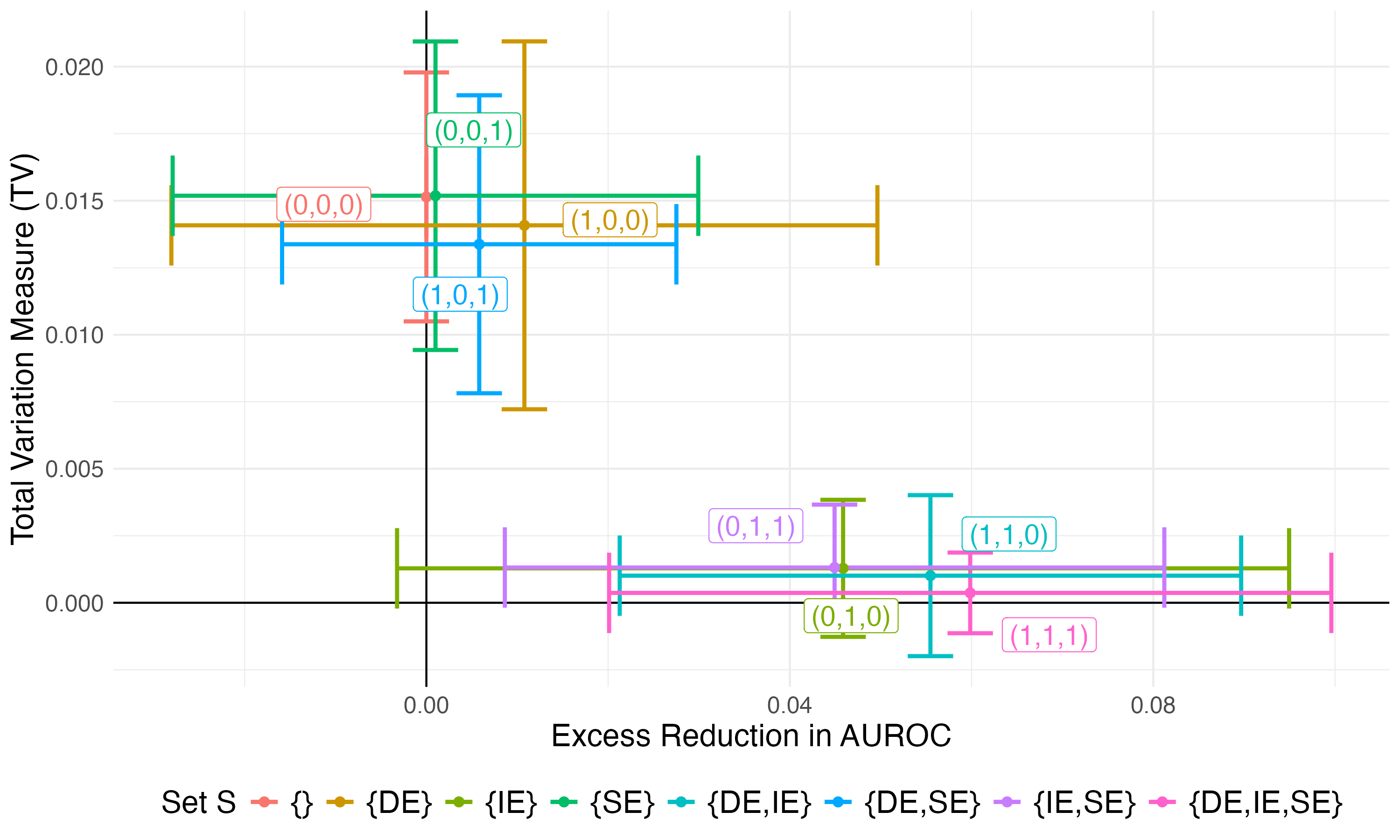} 
    \caption{Fairness/Utility of $\widehat Y^S$ predictors.}
    \label{fig:pareto-credit}
\end{subfigure}
\hfill
\begin{subfigure}[b]{0.45\textwidth}
    \centering
    \scalebox{0.85}{
    \begin{tikzpicture}[node distance=2.5cm]

\node (empty) {$\emptyset$};
\node[right of=empty, xshift=0.5cm] (I) {$\{I\}$};
\node[above of=I] (D) {$\{D\}$};
\node[below of=I] (S) {$\{S\}$};
\node[right of=D] (DI) {$\{D, I\}$};
\node[right of=S] (IS) {$\{I, S\}$};
\node[right of=I] (DS) {$\{D, S\}$};
\node[right of=DS, xshift=0.5cm] (DIS) {$\{D, I, S\}$};

\draw[->] (empty) -- (D) node[sloped, above, pos=\posmid] {\eTOd};
\draw[->] (empty) -- (I) node[sloped, above, pos=\posmid] {\eTOi};
\draw[->] (empty) -- (S) node[sloped, above, pos=\posmid] {\eTOs};

\draw[->] (S) -- (IS) node[sloped, above, pos=\posmid] {\sTOis};
\draw[->] (D) -- (DI) node[sloped, above, pos=\posmid] {\dTOdi};

\draw[->] (I) -- (DI) node[sloped, above, pos=\pospart] {\iTOdi};
\draw[->] (I) -- (IS) node[sloped, above, pos=\pospart] {\iTOis};
\draw[->] (D) -- (DS) node[sloped, above, pos=\pospart] {\dTOds};
\draw[->] (S) -- (DS) node[sloped, above, pos=\pospart] {\sTOds};

\draw[->] (DI) -- (DIS) node[sloped, above, pos=\posmid] {\diTOdis};
\draw[->] (IS) -- (DIS) node[sloped, above, pos=\posmid] {\isTOdis};
\draw[->] (DS) -- (DIS) node[sloped, above, pos=\posmid] {\dsTOdis};

\end{tikzpicture}
    }
    \caption{$\gpsel$ with PSEL (blue) and TVD values (red).}
    \label{fig:gpsel-credit}
\end{subfigure}
\caption{Application of Alg.~\ref{algo:pspl-attribution} on the UCI Credit dataset.}
\label{fig:credit-alg-1}
\end{figure*}
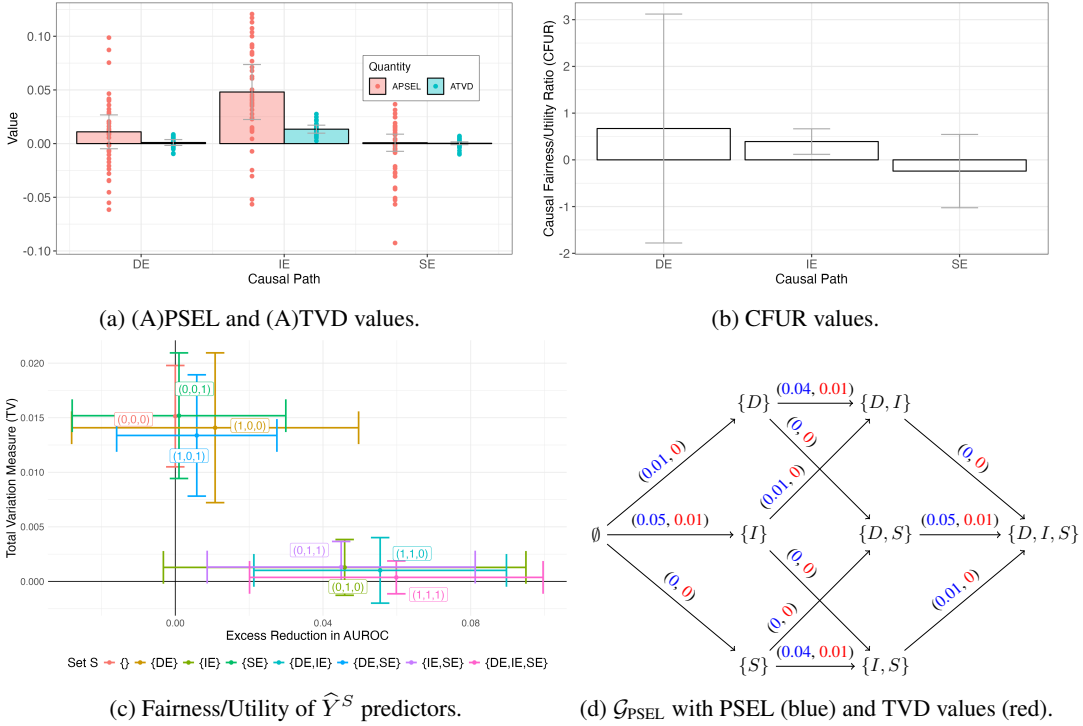
\begin{example}[UCI Credit \citep{uci2016credit}] \label{ex:credit}
    \upshape
    A commercial bank is developing an automated payment default scoring system based on the UCI Credit Dataset, collected in 2018. This dataset includes demographic variables $Z$ ($Z_1$ for age), the protected attribute gender $X$ ($x_0$ female, $x_1$ male), marital status $M$, education level $L$, employment status $E$, and financial history $F$. The institution aims to predict the binary outcome $Y$ that indicates whether the individual defaulted on a credit payment ($Y = 1$ for defaulting, $Y = 0$ otherwise). The grouping $\{X = {X}, Z = \{Z_1\}, W = \{L, M, E, F\}, Y = Y \}$ induces the standard fairness model \citep{plecko2022causal} for this analysis.

    The development team is focused on ensuring fairness in the risk of default predictions. 
    They examine the causal effects from the protected attribute $X$ to the predicted default probability $\widehat Y^S$, evaluating how much scores need to be adjusted to mitigate specific causal pathways, using the complement of AUROC ($1-$AUROC) as the loss metric. They employ Alg.\ref{algo:pspl-attribution} for this purpose and develop fair scoring models $\widehat Y^S$ using different $S$-sets as per Alg.\ref{algo:lagrange-training}.

    The analysis results are depicted in Fig.\ref{fig:credit-alg-1}, with uncertainty bars indicating standard deviations over 10 bootstrap repetitions. Fig.\ref{fig:apsel-credit}) shows the TVD and PSEL values. The team notes that for direct and spurious effects the APSEL and ATVD values are not significantly different from $0$. In words, they find that imposing a fairness constraint along these causal pathways does not significantly reduce the predictive power (in terms of reducing AUROC) of the predictor $\widehat Y ^ S$. 
    For the indirect effect, they find an APSEL value of 5\% of AUROC, and an ATVD of 1.5\%, showing that removing the indirect effect may reduce discrimination.

    The team then looks at CFUR values, finding that DE and SE values within one standard deviation contain $0$, again compatible no trade-off between fairness and accuracy. This is because the causal effects along DE and SE paths are already very close to $0$, and constraining such pathways has little effect on the predictor $\widehat Y ^ S$.
    The team then plots the TV values and the excess reduction in AUROC in Fig.~\ref{fig:pareto-credit}, for all predictors $\widehat Y ^S$ for the set $S$ ranging in the powerset of $\{ DE, IE, SE \}$. 
    The figure illustrates that the predictors $\widehat Y ^S$ are statistically indistinguishable in performance for two groups, namely:
    \begin{align}
        \widehat Y^{\emptyset}, \widehat Y^{\{\text{D}\}}, \widehat Y^{\{\text{S}\}}, \widehat Y^{\{\text{D, S}\}}, \text{ and} \\
        \widehat Y^{\{ I \}}, \widehat Y^{\{\text{D, I}\}}, \widehat Y^{\{\text{S, I}\}}, \widehat Y^{\{\text{D, I, S}\}}.
    \end{align}
    Once again, the removal of direct and spurious effects has little effect on the predictor, while the removal of the indirect effect seems to reduce discrimination.
    Based on these findings, the team considers whether to remove the indirect effect from the predictions. However, after a detailed discussion with a regulatory consultant, they decide that the usage of mediators $W$ is justifiable since $W$ reflects the spending behavior of the customers. Therefore, the team decides to use the $\widehat Y ^ \emptyset$ predictor.
\end{example}

\section{Path-Specific Fairness-Accuracy Trade-Offs} \label{appendix:path-specific}

In this appendix, we translate the results from the main text to a more general setting with path-specific causal effects. In particular, we consider a setting in which $m$ distinct paths are considered, labeled $S_1, \dots, S_m$. A particular choice we may consider is to set $S_1$ to measure the direct effect, $S_2$ to $S_{k+1}$ to measure the contributions of each of the confounders $Z_1, \dots, Z_k$ to the overall spurious effect, and $S_{k+2}$ to $S_{k + \ell+1}$ to measure the contributions of each of the mediators $W_1, \dots, W_\ell$ to the overall indirect effect. For explicit expressions for $S_{1}, \dots, S_m$ we refer the reader to a detailed discussion in \citep[Sec.~6]{plecko2024causal}, where some options for instantiating the effects $S_1, \dots, S_m$ are introduced and explained. 

We now go over the key results of the main paper that need to be adapted, and explain how they change when we consider path-specific effects.
A causally-fair predictor for a subset $S \subseteq \{ S_1, \dots, S_m \}$ of the above measures is defined as the following solution:
\begin{definition}[Path-Specific Causally Fair Predictor \citep{plecko2024reconciling}] \label{def:cfair-ps-predictor}
    The causally $S$-fair predictor $\widehat Y ^ S$ with respect to a loss function $L$ and pathways in $S$ is the solution to the following optimization problem:
    \begin{alignat}{2}
    \label{eq:inproc-causal-ps-objective}
    &\argmin_{f} &\ex \; L(Y, f(X, Z, W)) \\
    &\quad\text{s.t.} & S_i (f) = S_i (y) \mathbb{1}(S_i \notin S) \quad \forall i.
\end{alignat}
\end{definition}
The above definition is a generalization of Def.~\ref{def:cfair-predictor}. Building on this definition, we can state a more general version of Thm.~\ref{thm:tpl-decomposition}:
\begin{theorem}[Total Excess Loss Path-Specific Decomposition] \label{thm:tpl-ps-decomposition}
    The total excess loss $\text{PSEL} (\emptyset \to \{S_1, \dots, S_m\})$ can be decomposed into a sum of path-specific excess losses as follows:
    \begin{align}
        \text{PSEL} (\emptyset \to \{S_1, \dots, S_m\}) &= \sum_{i=0}^{m-1} \text{PSEL} (S_{[i]} \to S_{[i+1]}),
    \end{align}
    where $S_{[i]}$ is the set $S_1, \dots, S_i$, and $S_{[0]} = \emptyset$. 
\end{theorem}
For Algs.~\ref{algo:pspl-attribution} and ~\ref{algo:lagrange-training}, we also need some modifications. Crucially, in Alg.~\ref{algo:pspl-attribution}, we compute the APSEL$(S_i)$ values as
\begin{align}
         \frac{1}{m!} \sum_{\text{permutations } \sigma} \text{PSEL}(\tilde S_{[\sigma^{-1}(i)]}  \to \tilde S_{[\sigma^{-1}(i) + 1]} ),\label{eq:apsel-ps-compute}
    \end{align}
    and the $\text{ATVD}(S_i)$ value as
    \begin{align}
         \frac{1}{m!} \sum_{\text{permutations } \sigma} \text{TVD}(\tilde S_{[\sigma^{-1}(i)]}  \to \tilde S_{[\sigma^{-1}(i) + 1]} ), \label{eq:atvd-ps-compute}
    \end{align}
where $\sigma^{-1}(i)$ represents the position of index $i$ in the permutation, and $\tilde S$ is the permuted ordering of $S_1, \dots, S_m$ according to $\sigma$. The above summations range over all $m!$ permutations $\sigma$. 

Now, for adapting Alg.~\ref{algo:lagrange-training}, we first provide a definition of a path-specific Lagrangian causally-fair predictor:
\begin{definition}[Path-Specific Causal Lagrange Predictor $\widehat Y ^ S$] \label{def:ps-lagrange-form}
    The causally $S$-fair $\lambda$-optimal predictor $\widehat Y ^ S(\lambda)$ with respect to pathways in $S \subseteq \{ S_1, \dots, S_m \}$ and the loss function $L$ is the solution to the following optimization problem:
    {
    \small
    \begin{align} \label{eq:lagrange-ps-obj-1}
        \argmin_{f} & \; \ex \; L(Y, f(X, Z, W)) + 
        \\& 
        \lambda \sum_{i=1}^{m} \Big( S_i (f) - S_i (y) \mathbb{1}(S_i \notin S) \Big)^2 \label{eq:lagrange-ps-obj-last} 
    \end{align}
    }
\end{definition}
This definition builds on the notion introduced in Def.~\ref{def:cfair-predictor}, but considers a possibly broader range of causal pathways.
We also present Alg.~\ref{algo:lagrange-training-ps}, which is an appropriate generalization of Alg.~\ref{algo:lagrange-training} for the more general path-specific case.
\begin{algorithm}[t]
\caption{Causally-Fair Constrained Learning (CFCL)}
\label{algo:lagrange-training-ps}
\DontPrintSemicolon
 \KwIn{training data $\mathcal{D}_t$, evaluation data $\mathcal{D}_e$, set $S$, precision $\epsilon$}
Set $\lambda_{\text{low}} = 0$, $\lambda_{\text{high}} =$ large \;
\While{$|\lambda_{\text{high}} - \lambda_{\text{low}}| > \epsilon$}{
    set $\lambda_{\text{mid}} = \frac{1}{2} (\lambda_{\text{low}} + \lambda_{\text{high}})$\;
    fit a neural network to solves the optimization problem in Eqs.~\ref{eq:lagrange-ps-obj-1}-\ref{eq:lagrange-ps-obj-last} with $\lambda = \lambda_{\text{mid}}$ on $\mathcal{D}_t$ to obtain the predictor $\widehat Y ^ S (\lambda_{\text{mid}})$ \;
    compute the causal measures of fairness $S_1, \dots, S_m$ for $\widehat Y ^ S (\lambda_{\text{mid}})$ on evaluation data $\mathcal{D}_e$ \;
    test the hypothesis\;
    \For{$i\in {1, \dots, m}$}{
    \begin{align} \label{eq:h0-alg-ce}
        \hspace{-5pt} H_0^{S_i}: S_i(\widehat y ^ S (\lambda_{\text{mid}})) &= S_i(y) \cdot \mathbb{1}(S_i \notin S)
    \end{align}}
    \textbf{if} \textit{any $H_0^{S_i}$ rejected} \textbf{then} $\lambda_{\text{low}} = \lambda_{\text{mid}}$ \textbf{else}  $\lambda_{\text{high}} = \lambda_{\text{mid}}$
}
\Return{predictor $\widehat Y ^ S (\lambda_{\text{mid}})$}
\end{algorithm}
In Alg.~\ref{algo:lagrange-training-ps}, $m$ causal pathways are considered, and we perform a larger set of hypothesis tests when determining the appropriate value of the tuning parameter $\lambda$. In particular, in each step of the binary search procedure in Alg.~\ref{algo:lagrange-training-ps}, we test all of the hypotheses:
\begin{align} \label{eq:h0-pi-i}
        H_0^{S_i}: S_i(\widehat y ^ S (\lambda_{\text{mid}})) &= S_i(y) \cdot \mathbb{1}(S_i \notin S)
\end{align}
on the evaluation data $\mathcal{D}_e$.  In this way, we verify if each of the effects $S_i$ takes an appropriate value on a held-out dataset, which ensures that the correct path-specific constraints hold on the held-out data.

Further, we note that the statement of Prop.~\ref{prop:pspl-shapley} remains the same, and the proof of the more general statement considering $m$ causal pathways is given in Appendix~\ref{appendix:theorem-proofs} (see proof of Prop.~\ref{prop:pspl-shapley}). With this, the full set of results in the main paper are described for the setting with path-specific effects $S_1, \dots, S_m$. 


\end{document}